\documentclass[12pt]{article}

\usepackage{amsmath,amsthm, amsfonts, amssymb, amsxtra, amsopn}
\usepackage{pgfplots}
\usetikzlibrary{calc}
\pgfplotsset{compat=1.13}
\usepackage{pgfplotstable}
\usepackage{graphicx,grffile}
\usepackage{multirow}
\usepackage{booktabs}
\usepackage{listings}
\usepackage{cmap}
\usepackage{colortbl}
\usepackage{adjustbox}
\usepackage{epsfig}
\usepackage{bold-extra}

\usepackage[tableposition=top,font=small,skip=5pt]{caption}
\usepackage{subcaption}
\usepackage{makecell} 

\pgfkeys{
    /pgf/number format/precision=2, 
    /pgf/number format/fixed zerofill=true }
    
\pgfplotstableset{
    /color cells/min/.initial=0,
    /color cells/max/.initial=1000,
    /color cells/textcolor/.initial=,
    color cells/.code={%
        \pgfqkeys{/color cells}{#1}%
        \pgfkeysalso{%
            postproc cell content/.code={%
                \begingroup
                \pgfkeysgetvalue{/pgfplots/table/@preprocessed cell content}\value
\ifx\value\empty
\endgroup
\else
                \pgfmathfloatparsenumber{\value}%
                \pgfmathfloattofixed{\pgfmathresult}%
                \let\value=\pgfmathresult
                \pgfplotscolormapaccess
                    [\pgfkeysvalueof{/color cells/min}:\pgfkeysvalueof{/color cells/max}]%
                    {\value}%
                    {\pgfkeysvalueof{/pgfplots/colormap name}}%
                \pgfkeysgetvalue{/pgfplots/table/@cell content}\typesetvalue
                \pgfkeysgetvalue{/color cells/textcolor}\textcolorvalue
                \toks0=\expandafter{\typesetvalue}%
                \xdef\temp{%
                    \noexpand\pgfkeysalso{%
                        @cell content={%
                            \noexpand\cellcolor[rgb]{\pgfmathresult}%
                            \noexpand\definecolor{mapped color}{rgb}{\pgfmathresult}%
                            \ifx\textcolorvalue\empty
                            \else
                                \noexpand\color{\textcolorvalue}%
                            \fi
                            \the\toks0 %
                        }%
                    }%
                }%
                \endgroup
                \temp
\fi
            }%
        }%
    }
}

\PassOptionsToPackage{hyphens}{url}

\usepackage{hyperref}
\hypersetup{colorlinks=true,linkcolor=black,citecolor=black,urlcolor=blue,filecolor=black}
\hypersetup{pdfpagemode=UseNone,pdfstartview=}

\usepackage{enumitem}
\setlist[itemize]{noitemsep, topsep=0pt}

\advance\oddsidemargin by -0.35in
\advance\textwidth by 0.7in

\advance\topmargin by -0.4in
\advance\textheight by 0.8in

\long\def\symbolfootnotetext[#1]#2{\begingroup%
\def\thefootnote{\fnsymbol{footnote}}\footnotetext[#1]{#2}\endgroup}

\title{Classifying World War~II Era Ciphers with Machine Learning}

\author{Brooke Dalton\footnotemark[1]\ \ \ 
Mark Stamp\footnotemark[1]\,\,\footnotemark[2]}

\begin{document}

\symbolfootnotetext[1]{Department of Computer Science, San Jose State University}
\symbolfootnotetext[2]{mark.stamp$@$sjsu.edu}

\maketitle

\abstract
We determine the accuracy with which machine learning and deep learning techniques can 
classify selected World War~II era ciphers when only ciphertext is available. The specific ciphers 
considered are \texttt{Enigma}, \texttt{M-209}, \texttt{Sigaba}, \texttt{Purple}, 
and \texttt{Typex}. We experiment with three classic machine learning models, namely,
Support Vector Machines (SVM), $k$-Nearest Neighbors ($k$-NN), and Random Forest (RF).
We also experiment with four deep learning neural network-based models:
Multi-Layer Perceptrons (MLP), Long Short-Term Memory (LSTM), Extreme Learning Machines (ELM), 
and Convolutional Neural Networks (CNN). Each model is trained on 
features consisting of histograms, digrams, 
and raw ciphertext letter sequences. Furthermore,
the classification problem is considered under four distinct scenarios: 
Fixed plaintext with fixed keys, random plaintext with fixed keys, fixed plaintext 
with random keys, and random plaintext with random keys. Under the most realistic scenario, 
given~1000 characters per ciphertext, we are able to distinguish the ciphers
with greater than~97\%\ accuracy. In addition, we consider the accuracy of a 
subset of the learning techniques as a function of the length of the ciphertext messages.
Somewhat surprisingly, our classic machine learning models perform at least
as well as our deep learning models. 
We also find that ciphers that are more similar in design
are somewhat more challenging to distinguish, 
but not as difficult as might be expected.

\section{Introduction}

Many cipher machines were created and utilized by various nations during World War~II. For example, 
the \texttt{Enigma} is a rotor-based cipher machine that was invented as a commercial product and was later modified 
and deployed by the German military. Other rotor-based cipher machines of the same era include 
the \texttt{M-209 Converter} (henceforth, simply referred to as \texttt{M-209}) and \texttt{Sigaba}, which were both
created by the United States, and the British-built \texttt{Typex}. The well-known \texttt{Purple} cipher machine that was
used by Japan in World War~II employed a switch-based system. Although these ciphers have all been 
broken~\cite{bures2022cracking,chang2014cryptanalysis,gillogly1995ciphertext,chan2007cryptanalysis}, 
and cryptography 
has improved significantly since World War~II, accurately classifying ciphertext generated by these machines  
is an interesting challenge. There has been previous research on classifying classic ciphers by applying 
machine learning
techniques~\cite{kopal2020ciphers,krishna2019classifying,leierzopf2021massive,leierzopf2021detection,NCID}, 
but as far as the authors are aware, 
to date no such work has focused on the WWII-era cipher machines considered in this paper. 

Our research examines whether machine learning and deep learning techniques can classify 
selected World War~II era 
ciphers when only ciphertext is provided. We consider the five ciphers mentioned above, 
namely, \texttt{Enigma}, \texttt{M-209}, \texttt{Purple}, \texttt{Sigaba}, 
and \texttt{Typex}. For each cipher, we train and test machine learning techniques based on a variety of features; 
specifically, the raw ciphertext letter sequence, letter histograms, and letter digram statistics. We experiment with three popular 
classic machine learning techniques, namely, Support Vector Machines (SVM), $k$-Nearest Neighbors ($k$-NN), 
and Random Forest (RF) models. Additionally, we experiment with four deep learning techniques, namely, 
Multi-Layer Perceptrons (MLP), Long Short-Term Memory (LSTM) models, Extreme Learning Machines (ELM), 
and Convolutional Neural Networks (CNN). We train and test models under four distinct scenarios,
and we find that in the most realistic case, given sufficient ciphertext,
we can identify the correct cipher machine with accuracy in excess of~97\%.
We also conduct experiments to determine the relationship between the ciphertext length
and the accuracy of our various models.

The remainder of this paper is organized as follows. Section~\ref{chap:background} considers relevant  
background topics, including an overview to the ciphers considered and an introduction 
to the machine learning algorithms that we use. In Section~\ref{chap:data}, we discuss our experimental setup, 
including how the dataset was created, along with feature extraction. Section~\ref{chap:results} contains our 
experimental results, and we provide context and discussion of these results. In Section~\ref{chap:conclusion}, 
we give our conclusions and discuss possible directions for future work.

\section{Background}\label{chap:background}

In this section, we briefly review the history of the ciphers used in our experiments, 
including an overview of their physical layout, as well as their cryptographically significant 
components. We also introduce the machine learning algorithms that we employ in our experiments.

\subsection{World War~II Ciphers}

In our experiments, we consider five World War~II era machine ciphers, namely,
\texttt{Enigma}, \texttt{M-209}, \texttt{Sigaba}, \texttt{Purple}, 
and \texttt{Typex}. World War~II was 
noteworthy for the application of cryptographic technology. The volume of data that was need for secure 
military communications was much greater than in previous conflicts, and hence machine encryption was
widely used for the first time. Before these machines, codebooks were often used for military purposes.
While well-designed codebooks can be reasonably secure, they are cumbersome and slow.

Two of the five ciphers that we consider were used by the Axis powers, while the remaining three 
ciphers were used by the Allies. Both the Axis and Allied powers had confidence in the security of
their cipher machines, yet there were major cryptanalytic successes during the war, especially
by the Allies, who gained invaluable intelligence from their ability to read Axis ciphers.

\subsubsection{\texttt{Enigma}}

\begin{figure}[!htb]
\centering
\includegraphics[width=56.5mm]{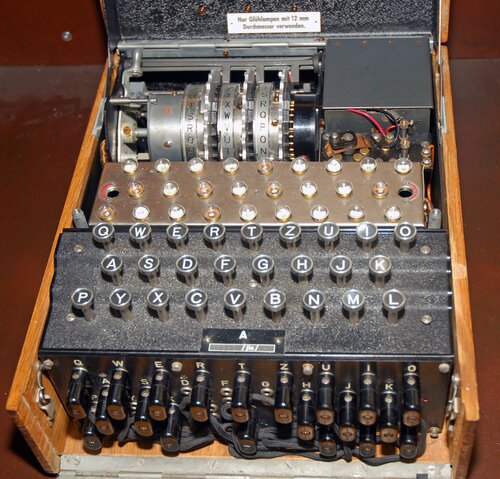}
\caption{\texttt{Enigma} cipher machine~\cite{tnmc}}\label{figure:Enigma}
\end{figure}

The \texttt{Enigma} cipher machine is a rotor-based cipher that was invented by German engineer Arthur Scherbius. 
During World War~II, the German military used modified versions of the original \texttt{Enigma} machine,
and  it became their primary military cipher system. In spite of clear indications that the
cipher had been compromised, it was used throughout the war. 
The version of the cipher machine that we consider appears in Figure \ref{figure:Enigma}.
This cipher has three rotors at the top, a keyboard for input, 
a lightboard to indicate the output, and a plugboard which is generally referred to by the German ``stecker.'' 
For our purposes, the key for this cipher consists of the initial rotor positions and stecker connections,
although various other internal setting can be changed.

A diagram illustrating the cryptographic elements of \texttt{Enigma} is given in Figure~\ref{fig:enigma}.
In this illustrated example, the letter~C is pressed on the keyboard, a stecker cable happens to 
map~C to~S, which is then passed through the stators (i.e., ``static rotors'' that do not rotate),
the three rotors, and the reflector, then back through the rotors and stators, yielding the letter~Z, 
which is mapped to~L by the stecker. Finally, the letter~L appears on the lightboard.

\begin{figure}[!htb]
  \centering
  \input figures/enigma.tex
  \caption{\texttt{Enigma} cipher diagram}\label{fig:enigma}
\end{figure}

With each keyboard letter that is pressed, the fast rotor~$F$ steps once, while the medium rotor~$M$ steps
once for each revolution of the fast rotor, and the slow rotor~$S$ steps once for each revolution of the medium 
rotor\footnote{There is one slight quirk to the rotor stepping described here. When a rotor steps, it causes the rotor to 
its right to also step. Since the rightmost rotor steps with each letter pressed, when the~$M$ rotor steps,
this quirk has no effect on the fast rotor. However, when the slow rotor steps, it causes the~$M$ rotor
to step, which results in a ``double stepping'' condition. Thus, the period of the rotor stepping 
is~$26\times 25\times 26\approx 2^{14.04}$, not~$26^3\approx 2^{14.10}$.}.

The crucial cryptographic components of the \texttt{Enigma} are the rotors, stators, and the stecker. 
Each rotor consists of a hardwired mapping between letters---when a rotor steps, it has the effect
of changing the overall permutation. Hence, the~\texttt{Enigma} is a polyalphabetic substitution,
with the current ``alphabet'' (i.e., permutation) determined by the configuration of the rotors.
Note that the \texttt{Enigma} is an electro-mechanical device, which implies that the reflector
permutation cannot have any fixed points, as a letter mapped to itself by the reflector would cause a 
short circuit. It is also interesting to note that the \texttt{Enigma} is self-inverse, and hence there is no
need for separate encryption and decryption modes.

\subsubsection{\texttt{M-209}}

\begin{figure}[!htb]
\centering
\includegraphics[width=55mm]{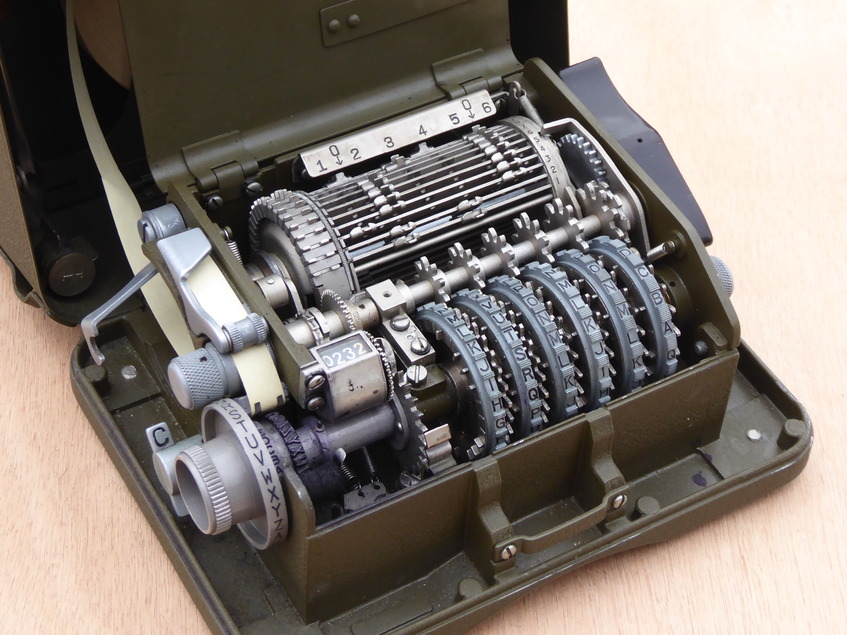}
\caption{\texttt{M-209} cipher machine~\cite{M209img}}
\label{figure:M-209}
\end{figure}

The \texttt{M-209} Converter, as shown in Figure \ref{figure:M-209}, 
is a mechanical rotor-based machine invented by Swedish engineer Boris Hagelin. 
This cipher was used by the United States military, and it is known as a ``lug and pin'' machine, for reasons
that will become clear momentarily.  The cipher includes six rotors, with the rotors from left to right
containing~26, 25, 23, 21, 19, and~17 letters, respectively. Each rotor letter has a small pin that can be set
to an active or inactive position. Behind the rotors is a cylindrical drum with~27 bars and two movable lugs per bar. 
There is an encoder wheel on the left of the machine with the alphabet on it---the alphabet also appears in reverse
order on a smaller wheel. When encrypting or decrypting, a handle on the right of the machine is turned, which causes
the drum lugs to interact with the rotor pins. In practice, the settings of the lugs and pins was a daily key,
while the message indicator (MI) consisted of the initial rotor settings.

A diagram illustrating the cryptographic components of \texttt{M-209} is given in Figure~\ref{fig:m209}.
To encrypt a letter, the encoder wheel is turned so that the plaintext letter appears at the indicated
position on the larger wheel. Then the handle is turned, which causes the encoder wheel to rotate,
with the ciphertext appearing on the smaller wheel, as well as being printed on the paper tape.
The cipher is its own inverse, so the decryption process is cryptographically identical to 
encryption.

\begin{figure}[!htb]
  \centering
  \input figures/m209.tex
  \caption{\texttt{M-209} cipher diagram}\label{fig:m209}
\end{figure}

In effect, at each encryption (or decryption) 
step, \texttt{M-209} generates a pseudo-random number in the range of~0 to~27, 
which determines the number of positions that the encoder wheel turns. The stepping is dependent on the settings
of the pins on the rotors and the lugs on the rotating drum---each drum bar for which a lug 
contacts an active pin causes the encoder wheel to rotate one position. After the pseudo-random step has been determined,
each rotor is rotated one position, exposing a new set of pins for the next 
letter encryption. The ciphertext (or plaintext) letters are printed on the paper tape, as well as being visible
on the smaller rotating letter wheel~\cite{Mark_tech}. 

As mentioned above, the \texttt{M-209} cipher is self-inverse, 
yet there is a switch that is set to specify encryption or decryption.
The difference between encryption and decryption modes is that for the former, ciphertext letters are printed in 
blocks of five on the paper tape, while for the latter, the letter~Z is rendered as word-space\footnote{The \texttt{M-209}
manual states that each word-space is to be encrypted as~Z. Thus, any plaintext~Z will decrypt as a space 
on the paper tape, and hence when decrypting, Z would be inferred from context.}.
Also, note that since the rotor lengths are relatively prime, the period until the pin positions are
certain to repeat is
$$
  26\times 25\times 23\times 21\times 19\times 17 \approx 2^{26.6} .
$$

\subsubsection{\texttt{Sigaba}}

\begin{figure}[!htb]
\centering
\includegraphics[width=55mm]{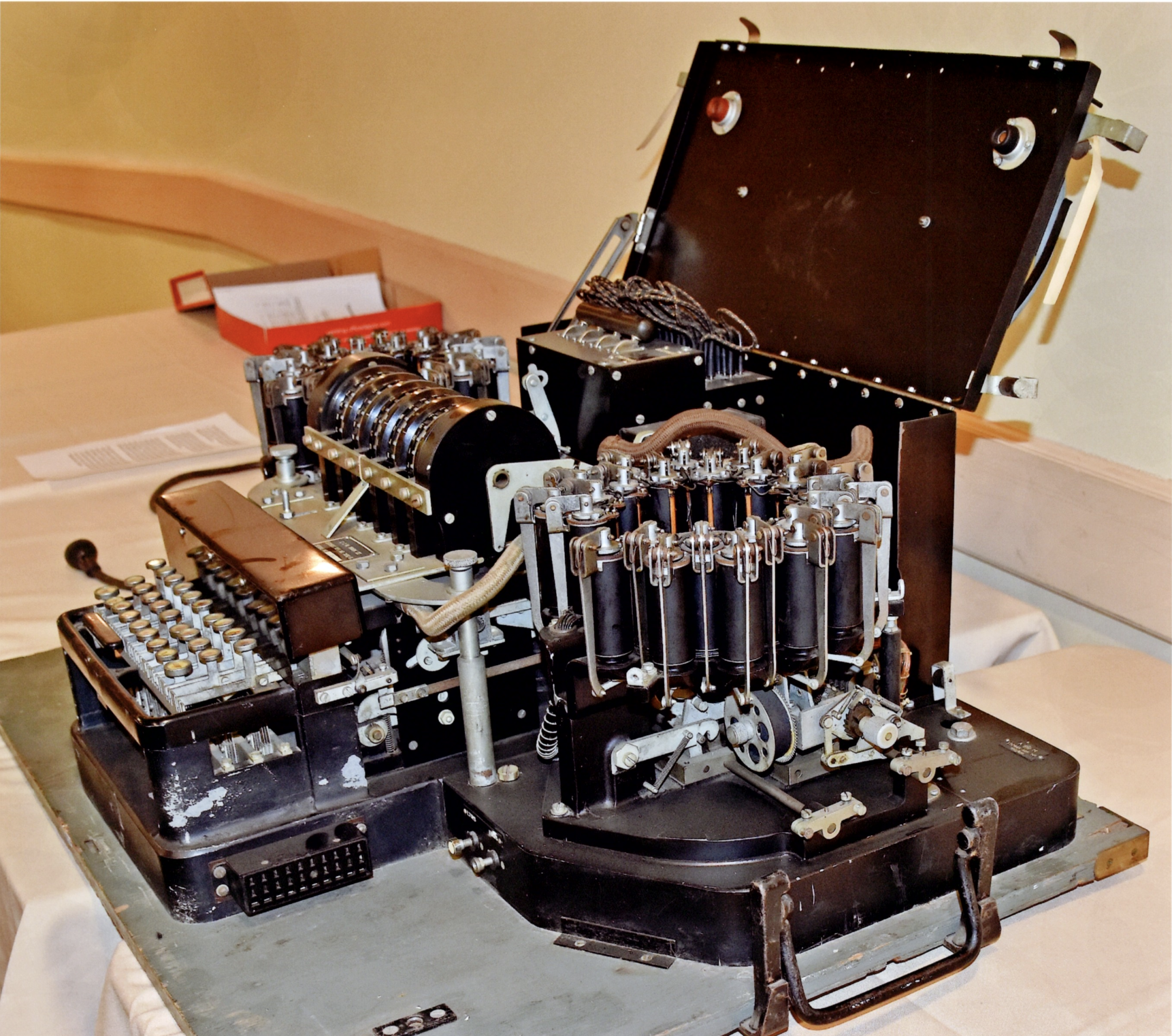}
\caption{\texttt{Sigaba} cipher machine~\cite{Sigabaimg}}
\label{figure:Sigaba}
\end{figure}

The \texttt{ECM Mark~II} cipher machine, better known as \texttt{Sigaba}, 
is pictured in Figure~\ref{figure:Sigaba}. \texttt{Sigaba}
is a rotor-based cipher created by American cryptographers,
and it was employed by the United States during World War~II and well into the~1950s. 
The cipher has five cipher cipher rotors, and separate banks 
of rotors that are used to determine the stepping of the cipher rotors. In effect, it is somewhat
analogous to an \texttt{Enigma} cipher with the another \texttt{Enigma} machine used to determine
the stepping of the cipher rotors.
No successful attacks on \texttt{Sigaba} are known to have occurred during 
its service lifetime. 

A diagram illustrating the cryptographically significant components of \texttt{Sigaba} 
is given in Figure~\ref{fig:sigaba}. The cipher machine includes a keyboard for entering plaintext (or ciphertex) letters, 
an output device for printing the ciphertext (or plaintext), and fifteen rotors. Five cipher rotors are used to encrypt and decrypt,
while the remaining ten rotors---five of which are referred to as control rotors and the remaining five are
known as index rotors---control the stepping of the five cipher rotors. With each letter that is encrypted
or decrypted, between one and four (inclusive) of the cipher rotors step.
This highly irregular stepping results in a far more complex polyalphabetic substitution cipher, 
as compared to rotor machines that use a regular stepping motion; see~\cite{stamp2007applied}
for more details on the inner workings of \texttt{Sigaba}.

\begin{figure}[!htb]
  \centering
  \input figures/sigaba.tex
  \caption{\texttt{Sigaba} cipher diagram}\label{fig:sigaba}
\end{figure}

The cipher and control rotors are interchangeable and reversible, which greatly increases the 
number of combinations available. The index rotors serve to permute the numbers~0 through~9,
and they do not step. The key consists of the order and orientation of the rotors and, 
of course, the initial position of each rotor can be adjusted.

\subsubsection{\texttt{Purple}}

The Japanese \texttt{Angooki Taipu B} cipher machine is better known by its nickname \texttt{Purple}, 
which is due to the binder color that United States cryptanalysts used to collect information on it. 
\texttt{Purple} was used for diplomatic communication during World War~II, and it was used to
encrypt the infamous ``14-part message'' that was sent from Tokyo to the Japanese embassy in Washington
the day before the Japanese attack on Pearl Harbor.

\begin{figure}[!htb]
\centering
\includegraphics[width=54mm]{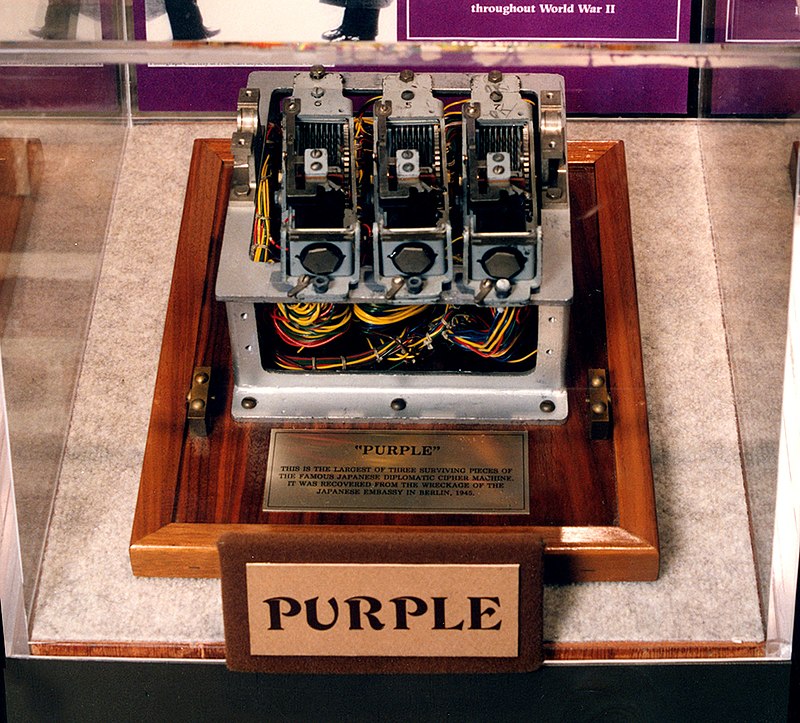}
\caption{Fragment of \texttt{Purple} machine~\cite{Purpleimg}}
\label{figure:Purple}
\end{figure}

No complete \texttt{Purple} cipher machine was ever found by the Allies although fragments were 
discovered after the war; Figure~\ref{figure:Purple} shows one such fragment.
Remarkably, American cryptographers were able to cryptanalyze the \texttt{Purple} cipher and regularly broke
encrypted messages---including reading the ``14-part message'' before the Japanese embassy in Washington
was able to decrypt it---in spite of never having seen the actual cipher machine. Perhaps even more remarkable,
the \texttt{Purple} analog that the American cryptanalysts constructed used the same telephone selector switches
that were in the actual \texttt{Purple} cipher.

The \texttt{Purple} cipher include an input plugboard, stepping switches, and an output 
plugboard~\cite{stamp2007applied}. Instead of using rotors to generate multiple alphabets, \texttt{Purple} has a number
of hardwired permutations, where the aforementioned telephone selector switches are used to switch between 
the permutations. There are~25 permutations per switch, resulting in a hardware implementation that is 
far more complex than a rotor-based machine, with the switches resembling a rats nest of wires.

A diagram illustrating the cryptographic components of \texttt{Purple} is given in Figure~\ref{fig:purple}.
When a letter is typed on the input keyboard, it is passed through a plugboard, similar to the
\texttt{Enigma} stecker, but including an unusual ``6-20 split'', whereby~6 letters of the alphabet are treated
differently than the other~20. If no cables are attached to the plugboard, the ``sixes'' are the vowels
and the ``twenties'' are the consonants. In any case, letters among the sixes are encrypted via a single switch, 
and also serve to determine the stepping of the three switches that yield the twenties permutation.

\begin{figure}[!htb]
  \centering
  \input figures/purple.tex
  \caption{\texttt{Purple} cipher diagram}\label{fig:purple}
\end{figure}

The 6-20 split is a major weakness of the \texttt{Purple} cipher, since the sixes have a different 
distribution from the twenties. We would expect that a machine learning algorithm will easily
distinguish \texttt{Purple} ciphertext from the rotor-based ciphers that we consider and, in fact,
we find that this is indeed the case.

\subsubsection{\texttt{Typex}}

\begin{figure}[!htb]
\centering
\includegraphics[width=56.5mm]{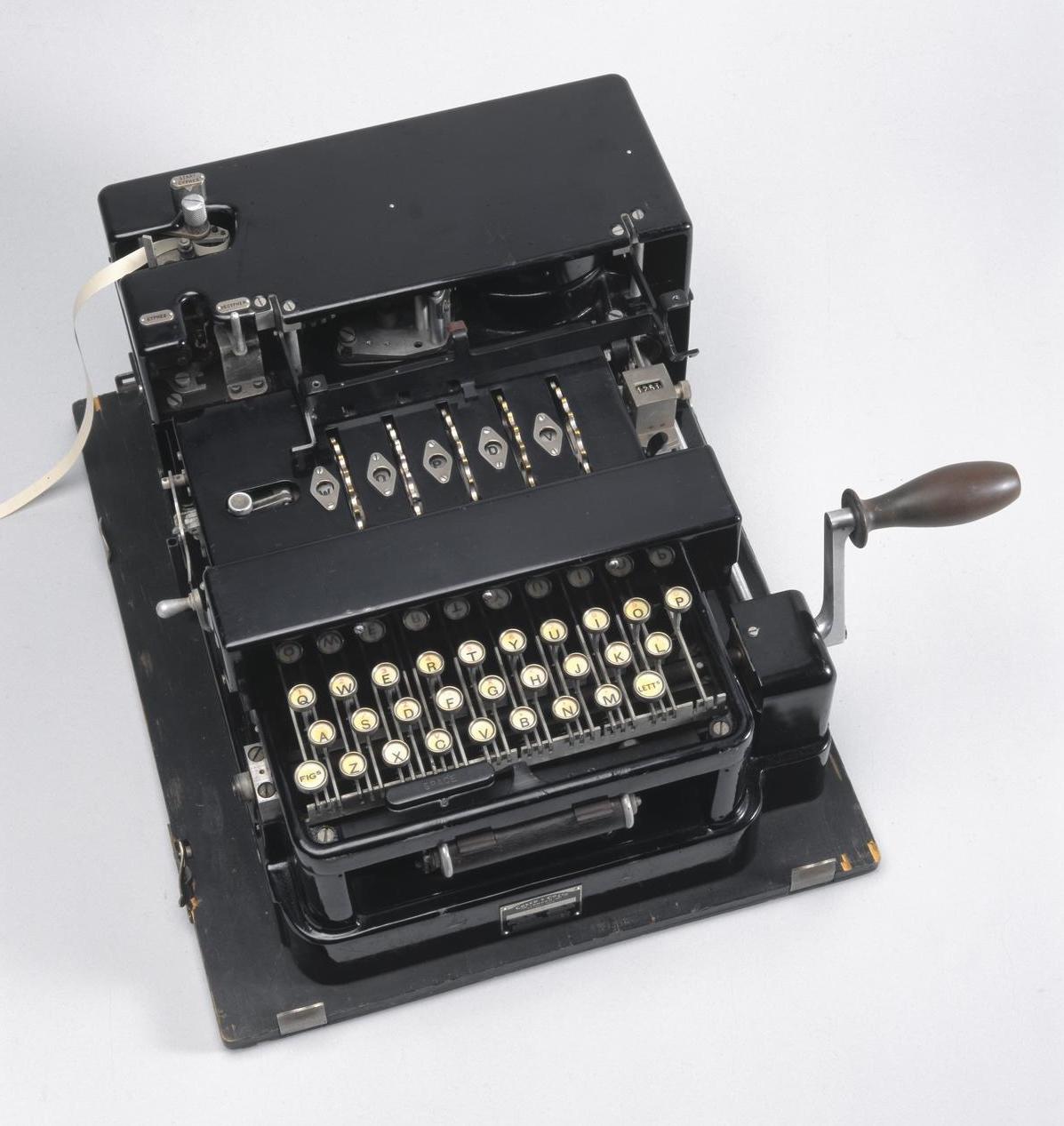}
\caption{\texttt{Typex} cipher machine~\cite{Typeximg}}
\label{figure:Typex}
\end{figure}

\texttt{Typex}, originally known as ``Enigma type with X-attachment'', 
is a rotor-based machine developed prior to WWII by \hbox{O.G.W.~Lywood} for the British 
government~\cite{erskine1997development}. As the original name implies, \texttt{Typex} 
was an adaptation of the commercial \texttt{Enigma},
with additional features designed to improve its security. There exist different variations of \texttt{Typex}, 
with a typical version having five rotors, a reflector, and a printer; later versions included an internal plugboard 
that was apparently used in place of the reflector~\cite{chang2014cryptanalysis}. 
A photo of a \texttt{Typex} machine appears in Figure~\ref{figure:Typex}.

A diagram illustrating the cryptographic components of \texttt{Typex} is given in Figure~\ref{fig:typex}.
Similar to the \texttt{Enigma} machine, when a letter is typed on the keyboard, it is sent through  
two stators, then the three stepping rotors, across the reflector, and back through the inverse of the same rotors
and stators. Unlike \texttt{Enigma}, the medium and slow rotors of \texttt{Typex} step more than once per revolution 
their neighboring rotor. This irregular and more frequent stepping pattern makes attacks more challenging,
as compared to \texttt{Enigma}.

\begin{figure}[!htb]
  \centering
  \input figures/typex.tex
  \caption{\texttt{Typex} cipher diagram}\label{fig:typex}
\end{figure}

\subsection{Learning Algorithms}\label{sect:learn}

In this section we introduce the classic machine learning and deep learning 
algorithms that we use in our experiments. These are popular 
learning techniques that have been shown to perform well in a wide variety of applications.

\subsubsection{Support Vector Machine}

Support Vector Machines (SVM) are one of the most popular supervised machine learning techniques 
for classification. The goal when training an SVM is to create a separating hyperplane where the
margin, i.e., the minimum distance from the hyperplane to training samples,
is maximized. However, this is not always possible to construct a separating hyperplane, 
as the data itself need not be linearly separable. 

When training an SVM,
the ``kernel trick'' enables us to map the data to a higher dimensional
space, where separating the classes is generally easier~\cite{stamp2022introduction}. 
In effect, the kernel trick allows us to embed a nonlinear transformation in
the process, without paying a significant penalty, in terms of computational efficiency. 

SVMs require a regularization hyperparameter~$C$ and, for
nonlinear kernels, $\gamma$. The precise definition of~$\gamma$ depends
on the specific nonlinear kernel function. The~$C$ parameter is related to the number of
misclassifications that are allowed when constructing separating hyperplanes.
The~$\gamma$ parameter can be viewed as defining how far the influence of each 
individual training sample extends. For our SVM experiments, we consider linear, 
Gaussian radial basis function (RBF), polynomial, and sigmoid kernel function.

\subsubsection{$k$-Nearest Neighbor}

The $k$-Nearest Neighbor ($k$-NN) algorithm is a machine learning technique 
for classification that is based on the distance between feature values. The $k$-NN
algorithm is said to be a ``lazy learner,'' since it requires no training---we simply 
classify a sample based on the~$k$ nearest samples in a specified training set.
In spite of its simplicity, as the size of the training dataset grows, $k$-NN tends
towards optimal, in a well-defined sense~\cite{stamp2022introduction}. 

\subsubsection{Random Forest}

Random Forest (RF) classifiers are ensembles of decision trees. RF uses a 
``divide and conquer'' approach to sample small subsets of the data and feature, 
with a decision tree constructed for each such subset. The RF classification is 
based on the combined predictions of its component decision trees~\cite{biau2016random}. 
Important hyperparameters in an RF include the number of estimators
(i.e., decision trees), maximum features (maximum number of features to sample in
any one decision tree), maximum depth of the decision trees, and ``criterion''
used to determine feature importance.

\subsubsection{Multilayer Perceptron}

The first of our deep learning algorithms in the Multilayer Perceptron (MLP),
which is sometimes known simply as an Artificial Neural Network (ANN) that consists of 
multiple layers of perceptrons. A perceptron is a mathematical abstraction of a 
neuron~\cite{lorrentz2015artificial}. An MLP is a feedforward technique that consists of 
at least three layers: the input layer, one or more hidden kayers, and an output layer. 

Important design decision for MLPa include the depth, that is, 
the number of hidden layers, the number of neurons per layer, the activation functions, 
and the objective function. The objective function is minimized to train the 
corresponding MLP~\cite{ruderSebastian}.

\subsubsection{Long Short-Term Memory}

Long Short-Term Memory (LSTM) networks are a class of Recurrent Neural Networks (RNNs) architecture. 
They can learn order dependencies in sequence prediction problems and are good at recognizing 
patterns in sequential data. RNNs, unlike other feedforward neural networks, have internal states 
that hold the context information about the previous inputs. Thus, the prediction at the current time 
step is dependent upon the context information from the previous time step. Plain ``vanilla'' RNNs 
struggle learning long term dependencies due to vanishing, exploding, or oscillating gradients
that occur during training, which limits their memory capacity. LSTMs mitigate gradient issues, 
allowing for larger time lags from when a feature is introduced to the time at which it is used 
by the model. Thus, LSTMs can utilize past information more effectively than 
generic RNN models~\cite{stamp2022introduction}. 

 \subsubsection{Extreme Learning Machines}

An Extreme Learning Machine (ELM) is a feedforward deep learning algorithm 
with a simple architecture. Typically, ELMs have an input layer, a single hidden layer, and an output layer, 
with the number of neurons at the hidden layer being a configurable hyperparameter~\cite{stamp2022introduction}. 
In an ELM, the hidden layer weights and biases are randomly assigned and are not
updated via training---only the output layer weights are trained, which can be
accomplished using a linear algebraic techniques. Thus, there is no need for the costly 
backpropagation algorithm that is used to train most other neural networking 
architectures. Consequently, ELMs are extremely efficient to train, although they typically
require more neurons than a comparable non-ELM architecture, which can make
them slightly less efficient for classification~\cite{HUANG2006489}.

\subsubsection{Convolutional Neural Network}

A Convolutional Neural Network (CNN) is a deep learning model that is designed to
efficiently deal with image classification. For images, fully-connected neural networking 
architectures are too costly to train, and they tend to overfit the training data.
CNNs are able to learn local dependencies, and they allow for a high degree
of translation invariance~\cite{stamp2022introduction}. Unlike other neural networking
classification techniques, CNNs learn convolutional filters. CNN models typically 
consist of an input layer, multiple convolutional layers, pooling layers, and a fully-connected 
output layer. The first convolutional layer is applied to the input data, and enables the
model to learn basic structures (e.g., edges), while each subsequent convolutional layer
is applied to the output of the previous layer, resulting in convolutions of convolutions,
which enables models to learn ever more abstract features. Pooling layers are used to 
reduce the dimensionality and increase translation independence. Many advanced 
CNN models have been pre-trained on vast image datasets---by retraining only the
output layer, such model can be successfully applied to problem domains that are
far different than the data on which they were originally trained.

\section{Dataset and Experimental Design}\label{chap:data}

This section details our dataset preparation and the design of our experiments. 
All models are developed in Python and use the \texttt{Scikit-learn} library. From 
the \texttt{Scikit-learn} library, the classification and metrics modules are used to 
evaluate the precision, recall, F1-score, and accuracy of the classifiers. 

\subsection{Dataset}

To generate the data required to classify these ciphers, CrypTool 2 (CT2), an application 
for cryptography and cryptanalysis, provided by the CrypTool project~\cite{CrypTool} 
has been 
used~\cite{kopal2018solving}. CT2 is an open source application that can be used 
to visualize, encrypt, and simulate most of the well-known Word War~II era cipher machines, 
along with many other ciphers. To generate each ciphertext message, plaintext from the 
Brown Corpus is used~\cite{francis1979brown}. The plaintext consists of~26 English alphabetic 
characters with no spaces, punctuation, or numbers. 
For each scenario, we use a balanced dataset of~1000 ciphertexts,
each consisting of~1000 characters, which gives us~1,000,000 characters of ciphertext
for each cipher. Since we consider five ciphers, we have a 
total of~5000 samples and~5,000,000 ciphertext characters.

\subsection{Feature Collection and Extraction}\label{sect:features}

Features play an important role in training any learning model. 
We consider three different features, which are described in this section.

\subsubsection{Histogram}

\begin{figure}[!htb]
\centering
\includegraphics[width=90mm]{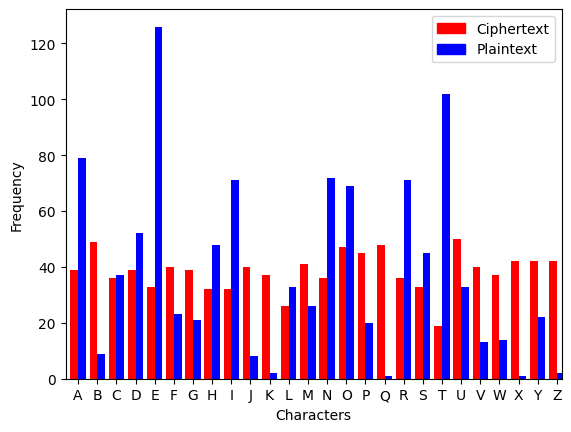}
\caption{Histogram example}
\label{figure:hist1}
\end{figure}

For our first set of experiments, we consider histograms based on monograph
statistics. That is, we simply count the frequency of each character per ciphertext message
and normalize by the number of characters. Figure \ref{figure:hist1} shows the frequency 
per character from one \texttt{Enigma} ciphertext sample and its corresponding plaintext.
Since \texttt{Enigma} is a polyalphabetic substitution cipher~\cite{stamp2007applied}, 
the plaintext letters are enciphered differently based upon their position in the text 
and hence they have a more uniform distribution, as compared to the plaintext.

\begin{figure}[!htb]
\centering
\includegraphics[width=100mm]{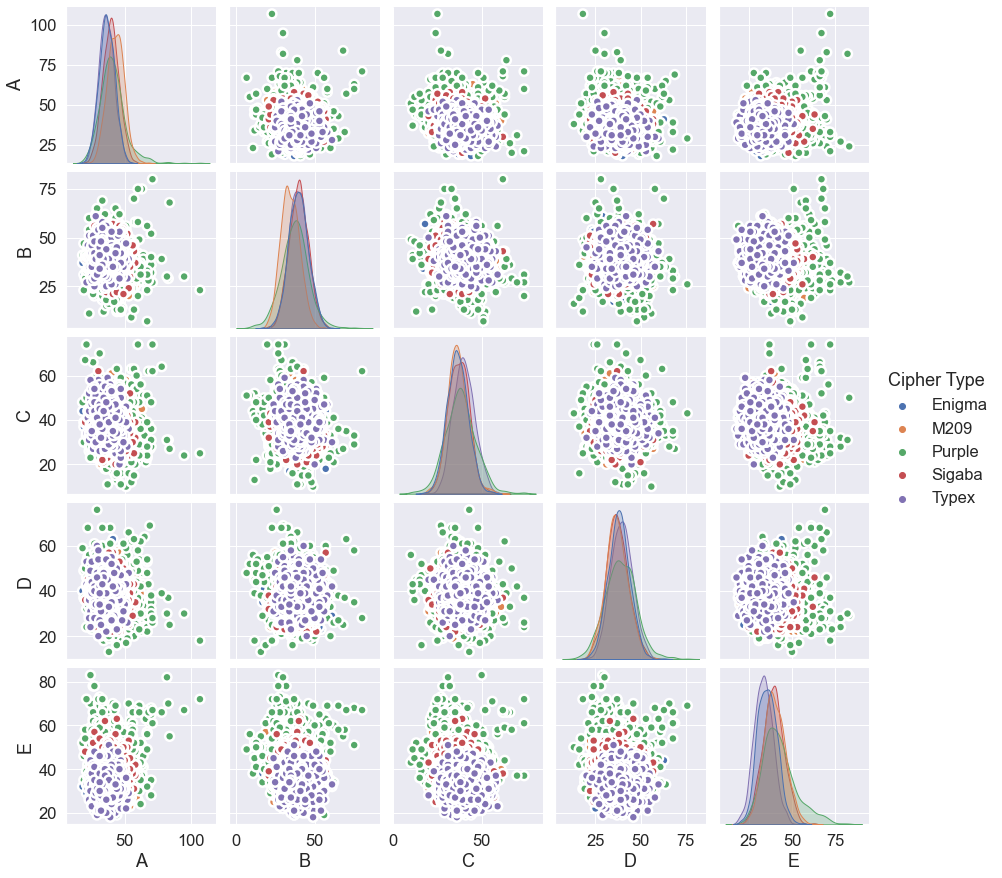}
\caption{Character, cipher, and frequency relationship}
\label{figure:hist2}
\end{figure}

Figure \ref{figure:hist2} shows the relationship between the first five letters of the alphabet 
that appear in ciphertext for each of the ciphers that we consider.
We observe that the \texttt{Purple} cipher clearly stands out from the rotor 
machines---especially with respect to the vowels---which is not surprising,
given the~6-20 split of alphabet used by \texttt{Purple}. These graphs  
indicate that the \texttt{Purple} cipher will be easy to distinguish from the
rotor cipher machines, but \texttt{Enigma}, \texttt{M-209}, \texttt{Sigaba}, and \texttt{Typex}
from each other may be far more challenging.

\subsubsection{Digram}

Frequency analysis can be extended to multiple characters, not just monograph statistics. 
In general, $n$-gram sequences consist of~$n$ consecutive characters.
In addition to the~$n=1$ case discussed above, we consider the~$n=2$ case, that
is, digraphs or digrams. Digram counts are analogous to single character counts, 
but instead of counting the occurrences of every character, we count the frequency 
of pairs of characters. Note that this can be viewed as a feature vector of
length~$26^2=676$. 

Figures~\ref{fig:digram}(a) and~(b) show heatmaps of digram
letter frequencies for English text and \texttt{Enigma} ciphertext, respectively.
We observe that the \texttt{Enigma} ciphertext has damped some of the roughness
that is inherent in the plaintext, but it is still far from uniform.

\begin{figure}[!htb]
\centering
\begin{tabular}{c}
\adjustbox{scale=0.385}{
\input figures/digramEnglish.tex
}
\\ \\[-2ex]
\adjustbox{scale=0.85}{(a) English plaintext}
\\ \\[-1ex]
\adjustbox{scale=0.385}{
\input figures/digramEnigma.tex
}
\\ \\[-2ex]
\adjustbox{scale=0.85}{(b) Enigma ciphertext}
\end{tabular}
\caption{Heatmaps of digram frequencies}\label{fig:digram}
\end{figure}

\subsubsection{Letter Sequence}

For our final set of classification experiments, we use the raw ciphertext letter sequences
as our feature vectors. Since each ciphertext sequence is of length~1000, this gives
us feature vectors of length~1000.

\subsection{Overview of Experiments}

Recall that for each of the five ciphers under consideration, we generate~1000 ciphertext
messages, each of length~1000. Once all of the ciphertexts have been created, the three types of
features discussed in Section~\ref{sect:features} are extracted for each ciphertext.
We then train each of the seven models discussed in Section~\ref{sect:learn}, 
giving us~21 models. Furthermore, we consider four different scenarios, as 
discussed below, which results in~84 distinct models.

\subsubsection{Fixed Plaintext and Fixed Keys}\label{sect:FF}

In this scenario, we first generate~1000 plaintext messages, then encrypt each using all 
five cipher types, using a single fixed key per cipher type. 
Once all~5000 ciphertext messages have been generated, the histogram
and digram feature vectors are generated, and the ciphertext messages
themselves require no additional processing for the letter sequence feature. 
We then train all seven models using each of these feature types.
We refer to this as the ``fixed-fixed'' scenario.

\subsubsection{Random Plaintext and Fixed Keys}\label{sect:RF}

For this scenario, we use a different plaintext for each encryption, 
and again use a single fixed key for each cipher type. Otherwise, the processing
of the data and generation of models is the same as in the fixed-fixed scenario.
We refer to this as the ``random-fixed'' scenario.

\subsubsection{Fixed Plaintext and Random Keys}\label{sect:FR}

Here, we use fixed plaintext, as in the fixed-fixed scenario, but for each encryption, 
we use a random key. Again, the processing
of the data and generation of models is the same as in the fixed-fixed scenario.
We refer to this as the ``fixed-random'' scenario.

\subsubsection{Random Plaintext and Random Keys}\label{sect:RR}

For our final scenario, we use a random plaintext and a random key for each encryption. 
The processing of the data and generation of models is the same as in the fixed-fixed 
scenario discussed above. We refer to this as the ``random-random'' scenario.

Note that this random-random scenario is the most realistic. Hence,
in the next section, we discuss our random-random experiments in detail,
while most of the results involving the other three scenarios are
relegated to the Appendix.

\section{Random-Random Scenario Experiments}\label{chap:results}

Since the random-random scenario is the most realistic, we provide a detailed discussion of our 
experiments and results for this case. Our experimental results for each of the other three 
scenarios is summarized in the Appendix.

For each of the three type of features considered, we generate seven models.
Each of these~21 model is trained and tested using an~80-20 split, that is,
80\%\ of the data is used for training, while the 
remaining 20\%\ of the data is used for testing.

To measure and compare the quality of our results, for each experiment 
we compute the accuracy, precision, recall, and F1-score.
Since there are~5000 samples for each experiment, and we use an~80-20
split, the total number of samples used for testing, or the support, is~1000.
Also, since we use a balanced split, the support for each cipher is~200.

For each model, a grid search is used to tune the hyperparameters over reasonable sets of values.
Table~\ref{tab:hyperA} lists the hyperparameters tested---as well as the hyperparameters selected---for 
each model when trained using each of the three feature types considered. 

\begin{table}[!htb]
    \centering
    \caption{Hyperparameters tested and selected}\label{tab:hyperA}
    \adjustbox{scale=0.665}{
    \begin{tabular}{c|ccccc} 
        \midrule\midrule
    Learning & \multirow{2}{*}{Hyperparameter} 
     & \multirow{2}{*}{Tested} & \multicolumn{3}{c}{Selected} \\ 
    		\cline{4-6} \\[-2.5ex]
    technique & & & Histogram & Digram & Sequence \\
    \midrule
    \multirow{3}{*}{SVM} & 
    $C$ & 1, 10, 100, 1000 & 100 & 1 & 1\\ 
    & $\gamma$ & 0.001, 0.0001 & 0.001 & 0.001 & 0.0001 \\ 
    & kernel & linear, poly, rbf, sigmoid & rbf & rbf & rbf\\
    \midrule 
    \multirow{3}{*}{$k$-NN} &
    Number of neighbors & (1,2,\ldots,200) & 83 & 98 & 74\\
    & Distance metric & Euclidean, manhattan, minkowski & Euclidean & Euclidean & manhattan\\
    & Weights & uniform, distance & uniform & uniform & uniform\\
    \midrule
    \multirow{3}{*}{RF} &
    Number of estimators & (1,2,\ldots,200) & 192 & 197 & 177\\ 
    & Max depth & 4, 5, 6, 7, 8 & 8 & 8 & 8\\
    & Criterion & gini, entropy & gini & gini & gini\\
    \midrule
    \multirow{5}{*}{MLP} &
    Activation function & tanh, relu & relu & relu & tanh\\
    & $\alpha$ & 0.0001, 0.05 & 0.0001 & 0.0001 & 0.0001 \\ 
    & Hidden layer size & (100,200,15), (150,100,50), (500,) & (500,) & (500,) & (500,)\\
    & Max iterations & 200, 500, 1000 & 200 & 200 & 200\\
    & Solver & sgd, adam & adam & adam & adam\\
    \midrule
    \multirow{2}{*}{ELM} &
    Activation function & relu, sigmoid, tanh & relu & tanh & tanh\\
    & Hidden neurons & (1,2,\ldots,1000) & 133 & 9696 & 995\\
    \midrule
    \multirow{3}{*}{LSTM} &
    Number of hidden layers & 1, 2, 3 & 3 & 2 & 2\\
    & Activation function & relu, tanh, softmax & softmax & softmax & softmax\\
    & Dropout rate & 0.1, 0.3, 0.9 & 0.3 & 0.3 & 0.3\\
    \midrule
    \multirow{3}{*}{CNN} &
    Number of layers & 3, 4, 5 & 5 & 5 & 5\\
    & Activation function & sigmoid, softmax & sigmoid & sigmoid & sigmoid\\
    & Dropout rate & 0.1, 0.3, 0.5 & 0.3 & 0.3 & 0.3\\
    \midrule\midrule    
    \end{tabular}
}
\end{table}


\subsection{Histogram Experiments}

In this section, we test each of the seven models discussed in Section~\ref{chap:background}
using the histogram features. Note that the histogram feature vector is of length~26 for each sample.
We provide a few relevant additional details on the training of each of the models. 
Finally, after discussing the models, we summarize the results for each.

When training our SVM models, we must determine the hyperparameter~$\gamma$,
the regularization parameter~$C$, and the kernel function. For the kernel function, 
we consider linear, polynomial, radial basis function, and sigmoid,
and we test reasonable values for~$\gamma$ and~$C$.
For the histogram features, we find that the sigmoid kernel yields the best results.

\begin{figure}[!htb]
\centering
\adjustbox{scale=0.65}{
\includegraphics{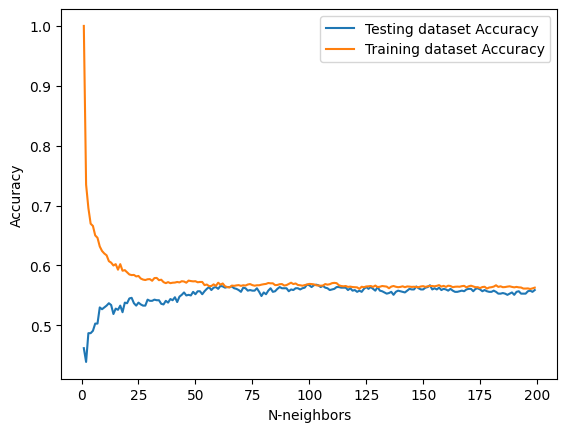}
}
\caption{Histogram $k$-NN number of neighbors and accuracy}\label{fig:k-NNHistBestN}
\end{figure}

For $k$-NN, we must first determine the value of~$k$, which is the number of nearest neighbors
considered when classifying a sample. In general, setting~$k$ too small results in overfitting,
while too large of a value can cause underfitting.
From the results in Figure~\ref{fig:k-NNHistBestN}, we 
determine that the optimal value is~$k=83$.

Another $k$-NN hyperparameter that we experimented with is selecting the best distance 
metric between neighbors. 
There are many different distance formulas available but we tested the most popular to reduce the 
hyperparameter grid computations. We also test whether the model does better when all points in each 
neighborhood are weighted equally or if they are weighted, based on their
distance from the sample being classified. 

For our Random Forest classifier, we experiment with the so-called number of ``estimators'', that is, the
number of decision trees used. We use ``sqrt'' for our maximum number of features, that is,
the square root of the total number of features in the dataset~\cite{scikit-learn}. 
The maximum depth, which represents the longest path between the root node and the leaf node,
was tested as part of our grid search.


For our MLP, we test a wide range of hyperparameters. Specifically, our grid search
includes the activation function, 
the regularization term, the size of the hidden layers, the maximum number iterations, 
and the weight optimization (or solver).


For our ELM, we experiment with the number of neurons in the hidden layer, and activation functions. 
It is important to change the encoding of our target variables with ELM, since the classification 
only works with discrete classes. We transform the single values of the target variable into a 
vector via one-hot encoding as given in Table~\ref{table:onehot}.

\begin{table}[!htb]
    \caption{One-hot encodings}\label{table:onehot}
    \centering
    \adjustbox{scale=0.85}{
    \begin{tabular}{ cc } 
    \midrule\midrule
    \texttt{Enigma} & (1, 0, 0, 0, 0)\\
    \texttt{M-209} & (0, 1, 0, 0, 0)\\
    \texttt{Purple} & (0, 0, 1, 0, 0)\\
    \texttt{Sigaba} & (0, 0, 0, 1, 0)\\
    \texttt{Typex} & (0, 0, 0, 0, 1)\\
    \midrule\midrule
    \end{tabular}
    }
\end{table}

\begin{figure}[!htb]
\centering
\adjustbox{scale=0.65}{
\includegraphics{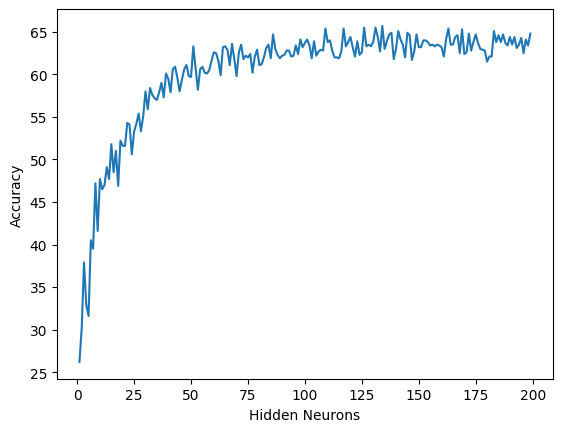}
}
\caption{Histogram ELM accuracy and number of hidden neurons}
\label{fig:histELMNeurons}
\end{figure}

From Figure~\ref{fig:histELMNeurons}, we see that
the accuracy of our ELM plateaus after about~50 hidden neurons.
However, the maximum accuracy is attained with~133 neurons, 
so that is what we use for this model.

\begin{figure}[!htb]
\centering
\begin{tabular}{ccc}
\includegraphics[width=50mm,height=39mm]{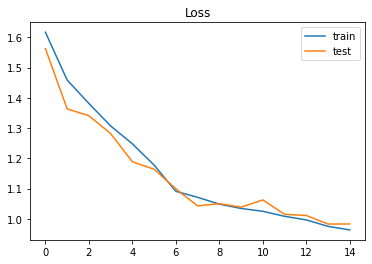}
& &
\includegraphics[width=50mm,height=39mm]{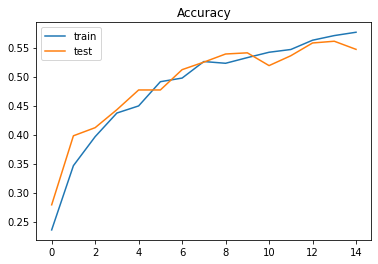}
\end{tabular}
\caption{Loss and accuracy of histogram LSTM model}\label{figure:lossAndAccHist}
\end{figure}

For our LSTM architecture, we experiment with the number of hidden layers, 
the activation function, and the dropout rate. A basic LSTM architecture 
includes an input layer, a single hidden layer, and an output layer. There are no clear 
guidelines on how to determine the number of layers for an LSTM, although~\cite{stathakis2009many} 
notes that, theoretically, one hidden layer often works well, and two layers are typically 
enough to learn more complex problems.
For the histogram features we found that two sequential layers, a dropout layer, 
and an output layer produce the best result.
From the loss and accuracy graphs in Figure~\ref{figure:lossAndAccHist},
we see the result of training for~15 epochs, and we note
that there is no indication of overfitting in these graphs.

We train a two-dimensional CNN. To create the ``images'' needed for training such a CNN, 
we use a heatmap of the histogram statistics. For our CNN architecture, we experiment with 
the number of convolutional layers, the activation function, and the dropout rate. For the 
histogram data we found that a CNN with three convolutional layers yields the best results. 

\begin{figure}[!htb]
\centering
\begin{tabular}{ccc}
\includegraphics[width=50mm,height=39mm]{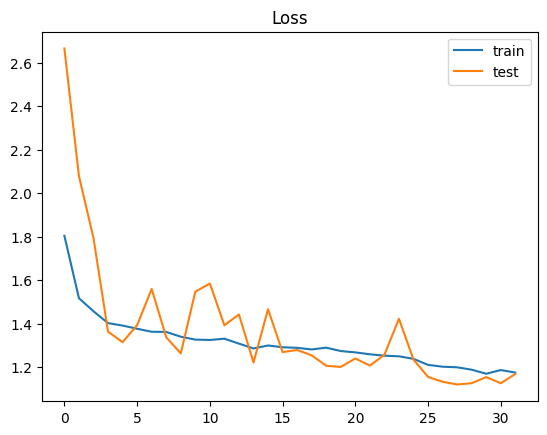}
& &
\includegraphics[width=50mm,height=39mm]{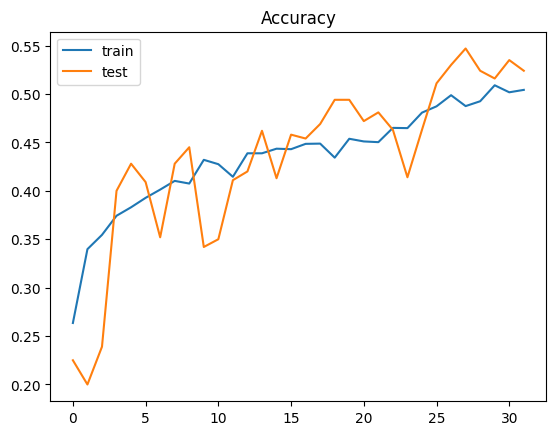}
\end{tabular}
\caption{Loss and accuracy of histogram CNN model}\label{figure:CNNlossAndAccHist}
\end{figure}

Based on the loss and accuracy graphs in Figure~\ref{figure:CNNlossAndAccHist}, 
we observe that after training for~32 epochs, the loss is about~1.07 and the accuracy is 
only about~0.49. We again see no evidence of overfitting in these graphs.

\begin{figure}[!htb]
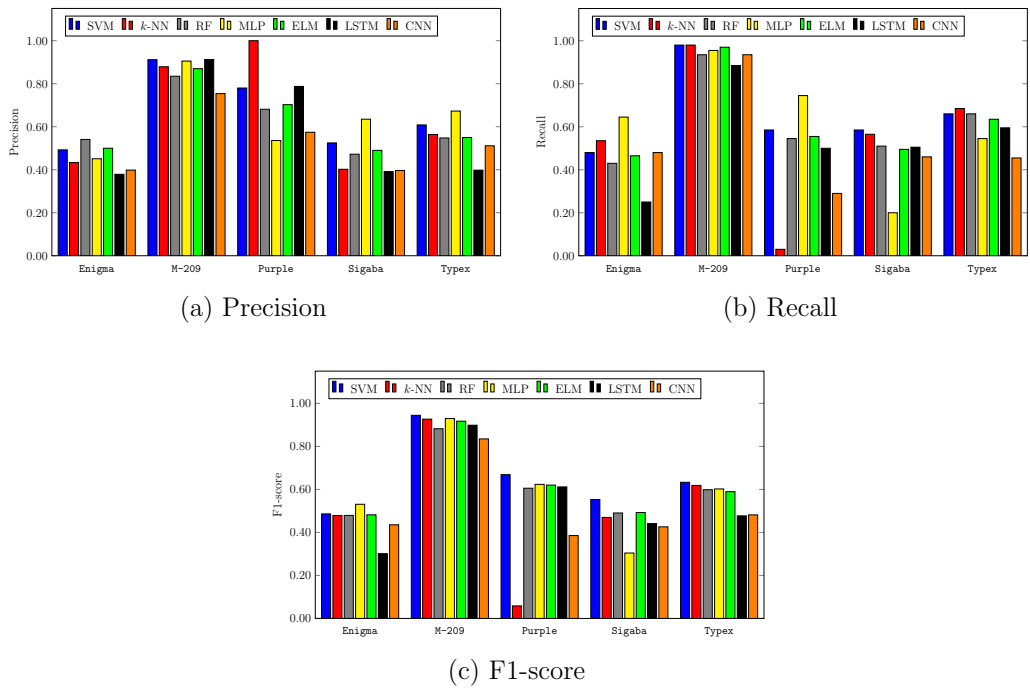

\centering
\begin{tabular}{cc}
\begin{adjustbox}{width=0.425\textwidth}
\input figures/random_random/histoPrec.tex
\end{adjustbox}
&
\begin{adjustbox}{width=0.425\textwidth}
\input figures/random_random/histoRecall.tex
\end{adjustbox}
\\
\adjustbox{scale=0.85}{(a) Precision}
&
\adjustbox{scale=0.85}{(b) Recall}
\\
\\
\multicolumn{2}{c}{
\begin{adjustbox}{width=0.425\textwidth}
\input figures/random_random/histoF1.tex
\end{adjustbox}
}
\\
\multicolumn{2}{c}{\adjustbox{scale=0.85}{(c) F1-score}}
\end{tabular} 
\caption{Precision, recall, and F1-scores for histogram features}\label{fig:histF1}
\end{figure}

Precision, recall, and F1-scores for each cipher and each learning technique---when 
trained on the histogram features as discussed above---are 
given in the bar graphs in Figure~\ref{fig:histF1}. We note that for this case,
\texttt{M-209} is easily distinguished from the rotor-based ciphers, while
the most similar ciphers---\texttt{Enigma}, \texttt{Sigaba}, 
and \texttt{Typex}---are not accurately distinguished.

A confusion matrix for each of the experiments in this section is given in 
the Appendix in Figure~\ref{figure:Hist}.
We observe that \texttt{Typex} and \texttt{Engima} are the most often confused with each other.
This is not surprising, given that \texttt{Typex} is a variant of the commercial version of \texttt{Engima}.


\subsection{Digram Experiments}

Training models using the digram features follows a similar pattern
as discussed above for the histogram features. and hence
we omit the details. As for the histogram features, we found no
evidence of overfitting for the models when using the digram features.
Recall that the hyperparameters selected in this case
are given in the penultimate column of Table~\ref{tab:hyperA}.

\begin{figure}[!htb]
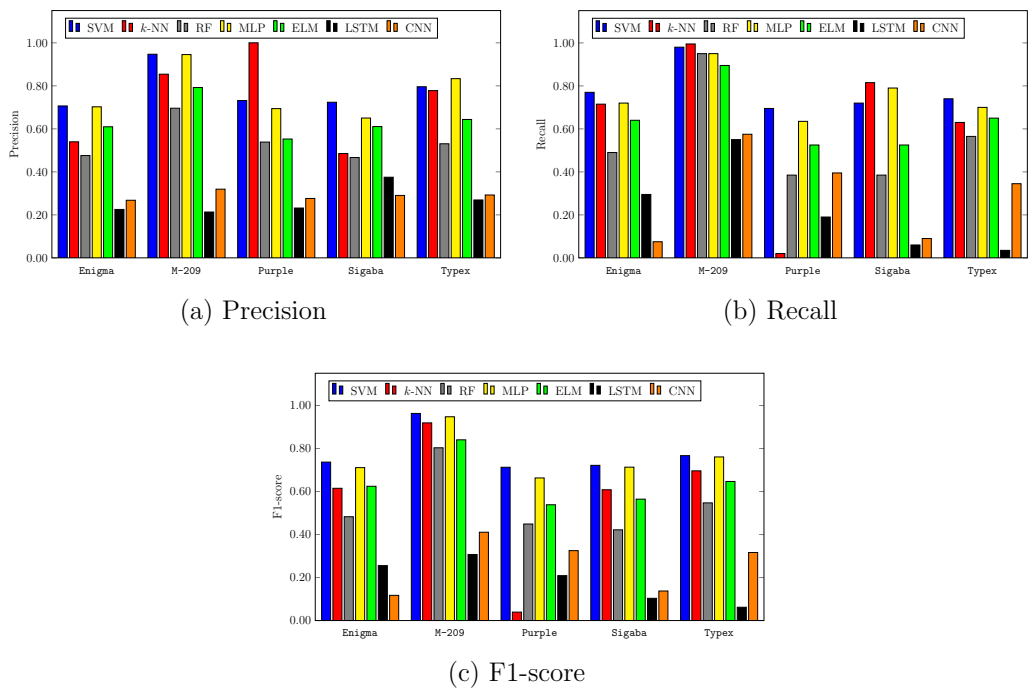

\centering
\begin{tabular}{cc}
\begin{adjustbox}{width=0.425\textwidth}
\input figures/random_random/digramPrec.tex
\end{adjustbox}
&
\begin{adjustbox}{width=0.425\textwidth}
\input figures/random_random/digramRecall.tex
\end{adjustbox}
\\
\adjustbox{scale=0.85}{(a) Precision}
&
\adjustbox{scale=0.85}{(b) Recall}
\\
\\
\multicolumn{2}{c}{
\begin{adjustbox}{width=0.425\textwidth}
\input figures/random_random/digramF1.tex
\end{adjustbox}
}
\\
\multicolumn{2}{c}{\adjustbox{scale=0.85}{(c) F1-score}}
\end{tabular} 
\caption{Precision, recall, and F1-scores for digram features}\label{fig:digramF1}
\end{figure}

The results of our digram experiments are summarized in Figure~\ref{fig:digramF1}.
We note that the results are similar to those obtained for the histogram
features, above.

Confusion matrices for all of the digram experiments in this section are given in 
the Appendix in Figure~\ref{figure:Digram}.


\subsection{Letter Sequence Experiments}

Our models trained using letter sequence features again
follow a similar pattern
as for the histogram and digram features, so
we omit the details. We again found no
evidence of overfitting for the models when using the letter sequence features.
The hyperparameters selected in this case
are given in final column in Table~\ref{tab:hyperA}.

\begin{figure}[!htb]
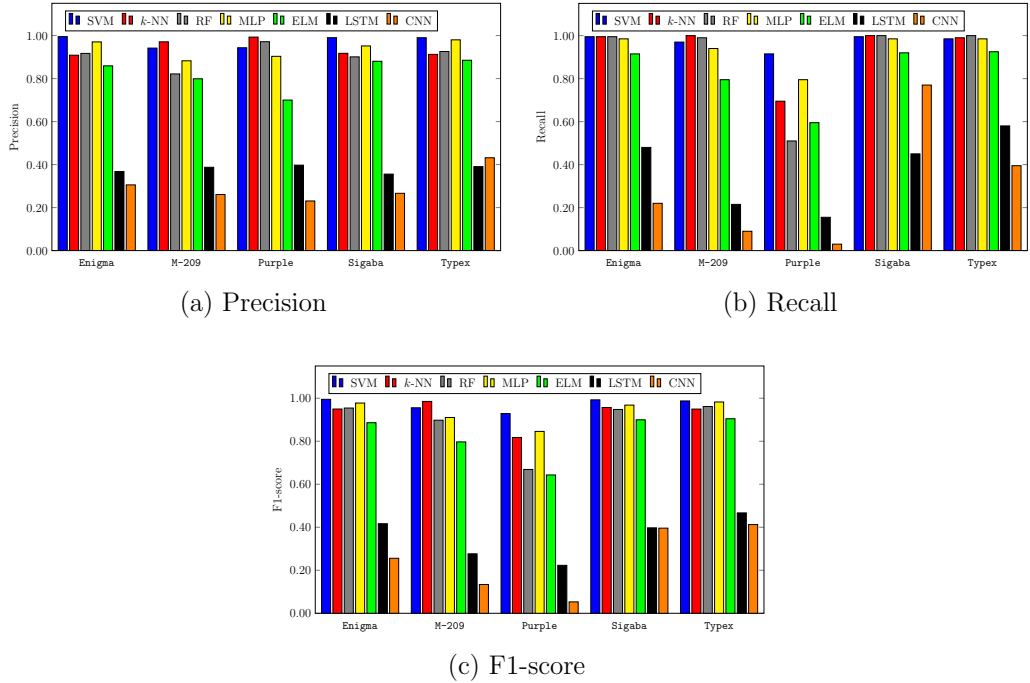

\centering
\begin{tabular}{cc}
\begin{adjustbox}{width=0.425\textwidth}
\input figures/random_random/seqPrec.tex
\end{adjustbox}
&
\begin{adjustbox}{width=0.425\textwidth}
\input figures/random_random/seqRecall.tex
\end{adjustbox}
\\
\adjustbox{scale=0.85}{(a) Precision}
&
\adjustbox{scale=0.85}{(b) Recall}
\\
\\
\multicolumn{2}{c}{
\begin{adjustbox}{width=0.425\textwidth}
\input figures/random_random/seqF1.tex
\end{adjustbox}
}
\\
\multicolumn{2}{c}{\adjustbox{scale=0.85}{(c) F1-score}}
\end{tabular} 
\caption{Precision, recall, and F1-scores for letter sequence features}\label{fig:seqF1}
\end{figure}

The results of our letter sequence experiments are summarized in Figure~\ref{fig:seqF1}.
We note that the best of these results are clearly much stronger than 
those obtained for the histogram or digram features, above.

Confusion matrices for all of the digram experiments in this section are given in 
the Appendix in Figure~\ref{figure:Seq}.


\subsection{Reduced Ciphertext Length}


All of the experiments above are based on ciphertext messages of length~1000. In this section, we
consider the effect on classification accuracy when less ciphertext is available. Since the random-random
case is the most realistic, and since SVM and MLP
models with letter sequence and diagram features performed best, this scenario with 
these models and features are considered here.

\begin{figure}[!htb]
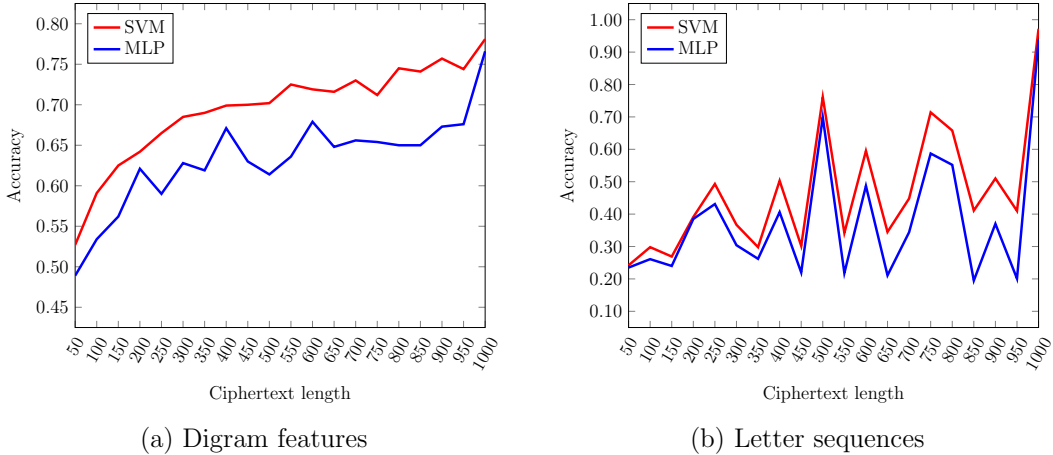

  \centering
  \begin{tabular}{cc}
  \adjustbox{scale=0.785}{
  \input figures/shortDigram.tex
  }
  &
  \adjustbox{scale=0.785}{
  \input figures/shortSeq.tex
  }
  \\
  \adjustbox{scale=0.85}{(a) Digram features}
  &
  \adjustbox{scale=0.85}{(b) Letter sequences}
  \end{tabular}
  \caption{Accuracy as a function of ciphertext length}\label{fig:short}
\end{figure}

In Figure~\ref{fig:short}(a), we give results for both the SVM and MLP models, based on the digram features.
We observe that the SVM performs better for every length tested. Furthermore, for the SVM model,
the accuracy drops from~0.781 for ciphertext of length~1000 to~0.685 for ciphertext of length~300,
and the model achieves a respectable accuracy of~0.527 with just~50 ciphertext characters.

Note that for the digram feature, regardless of the ciphertext length, the feature
vector is of length~$26^2 = 676$. Hence, we can---and do---use the models 
trained on~1000 ciphertext symbols to test each ciphertext length.

In Figure~\ref{fig:short}(b), we give results for both the SVM and MLP models, based on the ciphertext
letter sequences. In this case, the results are wildly inconsistent; for example, the MLP drops 
from an accuracy of~0.938 with ciphertext of length~1000 to an accuracy that is no better
than guessing (0.201, to be precise) for length~950, and then rebounds to an accuracy of~0.699 at length~500.



For each different ciphertext length, the models in Figure~\ref{fig:short}(b)
must be retrained, as the feature vector length has changed. Due to time constraints,
we did not tune the hyperparameters when retraining the models, but instead used the 
optimal hyperparameters for the length~1000 models. Consequently, the unstable results in
Figure~\ref{fig:short}(b) indicate that these models are likely very sensitive
to proper tuning of the hyperparameters, at least for the classic cipher classification 
problem considered in this paper.

\subsection{Discussion}

\begin{figure}[!htb]
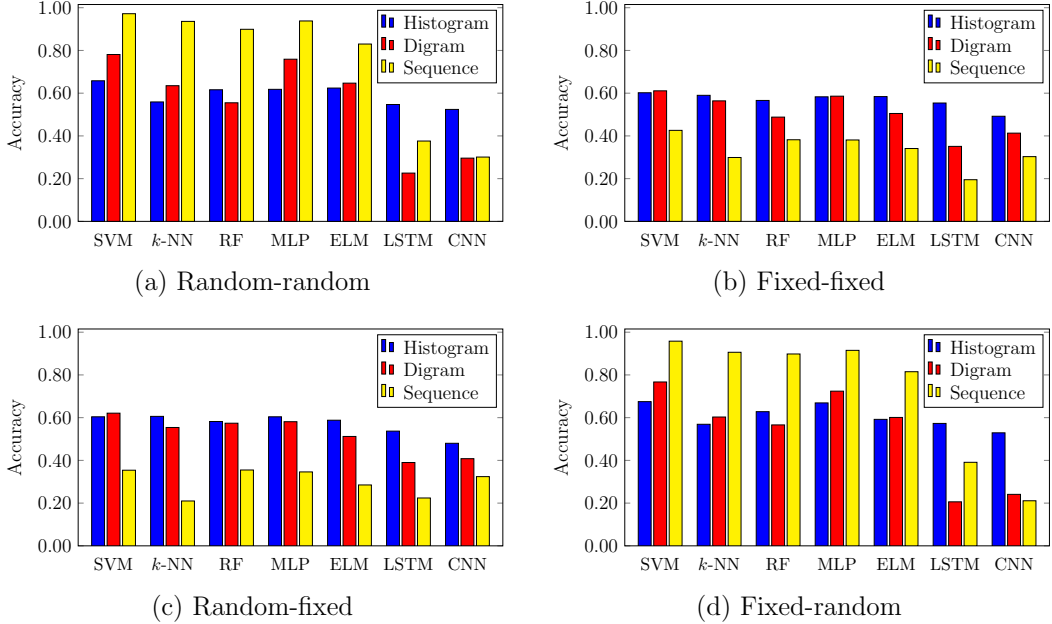

\centering
\begin{tabular}{cc}
\adjustbox{scale=0.65}{
\input figures/random_random/accuracy.tex
}
&
\adjustbox{scale=0.65}{
\input figures/fixed_fixed/accuracy.tex
}
\\
\adjustbox{scale=0.85}{(a) Random-random}
&
\adjustbox{scale=0.85}{(b) Fixed-fixed}
\\ \\[-1.5ex]
\adjustbox{scale=0.65}{
\input figures/random_fixed/accuracy.tex
}
&
\adjustbox{scale=0.65}{
\input figures/fixed_random/accuracy.tex
}
\\
\adjustbox{scale=0.85}{(c) Random-fixed}
&
\adjustbox{scale=0.85}{(d) Fixed-random}
\end{tabular}
\caption{Model accuracies for each scenario}\label{fig:modelacc}
\end{figure}

For our random-random scenario experiments,
the overall accuracy for each of the seven models and each of the three
feature types is given in Figure~\ref{fig:modelacc}(a). The corresponding
accuracies for the fixed-fixed, random-fixed, and fixed-random scenarios
are given in Figures~\ref{fig:modelacc}(b) through~(d), respectively. 

From the graphs  in Figure~\ref{fig:modelacc},
it is apparent that random keys are necessary for the models to learn to accurately distinguish
between the ciphers. Curiously, in the fixed key cases, the letter sequence feature
performs particularly poorly, while it is the best feature
in both of the random-key scenarios. 

Classifying ciphertexts using machine learning is a challenging task, as any well-designed cipher 
should produce ciphertext that is close to ``random.'' Nevertheless, in our random-random 
scenario experiments, the best classification accuracy that we obtained was slightly more 
than~97\% for an SVM model trained on ciphertext letter sequences, 
while MLP and $k$-NN models---also trained on letter 
sequences---were next best at~93.8\% and~93.6\% accuracy, respectively. 

In contrast to the letter sequence feature, for the histogram feature, 
our best accuracy for the random-random scenario was only~66\%, while for the digram feature,
the best accuracy was~78\%. For both of these feature types, the SVM was also the best model.
In all scenarios, the LSTM and CNN models perform relatively poorly.

\section{Conclusion and Future Work}\label{chap:conclusion}

For the five different WWII ciphers considered in this research, we have shown that 
machine learning algorithms can determine the cipher type from ciphertext only, 
with high accuracy.
Our best results for the most realistic scenario (i.e., random plaintext and random keys)
are summarized in Figure~\ref{fig:modelacc}(a), above.
Overall, the SVM performed best,
with MLP and $k$-NN being next best.

We found that among the five cipher machines considered,
\texttt{M-209} was consistently the easiest cipher to classify,
and \texttt{Purple} was generally relatively easy to distinguish. 
These results are not surprising, since the rotor system in \texttt{M-209}
is much different than that used in the other rotor-based ciphers considered, while
the~6-20 split used by \texttt{Purple} should make its switch-based system stand out. 
The most challenging ciphers to distinguish between are \texttt{Typex} and \texttt{Enigma} which, again,
is not surprising, since these ciphers share a common ancestor in the form of the original commercial 
\texttt{Enigma} machine. However, in spite of being the most challenging case, 
distinguishing between \texttt{Typex} and \texttt{Enigma} is fairly easy for
our best models.

Future work could involve additional ciphers, additional learning techniques, 
ensemble techniques, further hyperparameter tuning,
and consideration of additional features and feature engineering techniques. 
For example, our CNN results were not impressive, but using pre-trained models, 
such as VGG19 or any of the popular ResNet models
would be well worth considering. Also, we expected the LSTM to  
perform well on the letter sequence features, yet the results for this model 
were consistently poor, perhaps due to insufficient training. This too is worth further study.

\section{Acknowledgment}

The authors sincerely thank Nils Kopal for his help in generating the data that was essential
for the success of this project.

\bibliographystyle{plain}

\bibliography{references.bib}


\section*{Appendix}

In this appendix, we provide confusion matrices for the random plaintext
with random keys (random-random) scenario. 
In addition, we summarize the results for the other three scenarios
considered, namely, the fixed plaintext with fixed keys (fixed-fixed), 
random plaintext with fixed keys (random-fixed), and fixed plaintext with 
random keys (fixed-random).


\subsection*{Random-Random Scenario Confusion Matrices}

\begin{figure}[!htb]
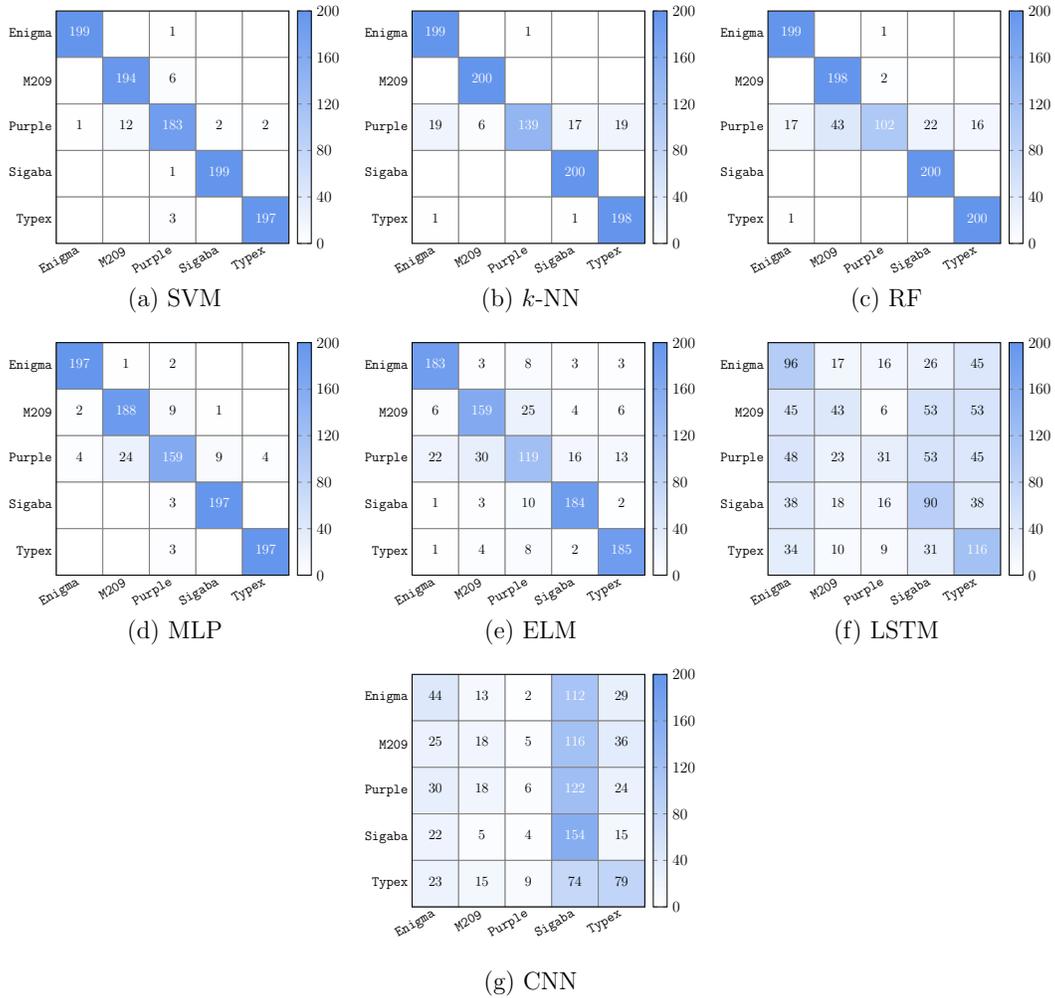

\centering
\adjustbox{scale=0.525}{
\begin{tabular}{ccc}
\input figures/random_random/conf_seq_SVM.tex
&
\input figures/random_random/conf_seq_kNN.tex
&
\input figures/random_random/conf_seq_RF.tex
\\
\adjustbox{scale=1.5}{(a) SVM}
&
\adjustbox{scale=1.5}{(b) $k$-NN}
&
\adjustbox{scale=1.5}{(c) RF}
\\ \\
\input figures/random_random/conf_seq_MLP.tex
&
\input figures/random_random/conf_seq_ELM.tex
&
\input figures/random_random/conf_seq_LSTM.tex
\\
\adjustbox{scale=1.5}{(d) MLP}
&
\adjustbox{scale=1.5}{(e) ELM}
&
\adjustbox{scale=1.5}{(f) LSTM}
\\ \\
& \input figures/random_random/conf_seq_CNN.tex &
\\ \\
& \adjustbox{scale=1.5}{(g) CNN}
\end{tabular}
}
\caption{Random-random confusion matrices for letter sequence feature}\label{figure:Seq}
\end{figure}

Here, we give confusion matrices for each of the 
random-random scenario experiments discussed in Section~\ref{chap:results}. 
Recall that we considered
seven models (SVM, $k$-NN, RF, MLP, ELM, LSTM, and CNN) 
and three feature types (histogram, digram, and letter sequence),
and hence we have a total of~21 confusion matrices.
The confusion matrices for the histogram features are in Figure~\ref{figure:Hist},
while those for the digram features are in Figure~\ref{figure:Digram},
and the confusion matrices for the letter sequence features
are in Figure~\ref{figure:Seq}.

\begin{figure}[!htb]
\centering
\adjustbox{scale=0.525}{
\begin{tabular}{ccc}
\input figures/random_random/conf_hist_SVM.tex
&
\input figures/random_random/conf_hist_kNN.tex
&
\input figures/random_random/conf_hist_RF.tex
\\
\adjustbox{scale=1.5}{(a) SVM}
&
\adjustbox{scale=1.5}{(b) $k$-NN}
&
\adjustbox{scale=1.5}{(c) RF}
\\ \\
\input figures/random_random/conf_hist_MLP.tex
&
\input figures/random_random/conf_hist_ELM.tex
&
\input figures/random_random/conf_hist_LSTM.tex
\\
\adjustbox{scale=1.5}{(d) MLP}
&
\adjustbox{scale=1.5}{(e) ELM}
&
\adjustbox{scale=1.5}{(f) LSTM}
\\ \\
& \input figures/random_random/conf_hist_CNN.tex &
\\ \\
& \adjustbox{scale=1.5}{(g) CNN}
\end{tabular}
}
\caption{Random-random confusion matrices for histogram features}\label{figure:Hist}
\end{figure}

\begin{figure}[!htb]
\centering
\adjustbox{scale=0.525}{
\begin{tabular}{ccc}
\input figures/random_random/conf_digram_SVM.tex
&
\input figures/random_random/conf_digram_kNN.tex
&
\input figures/random_random/conf_digram_RF.tex
\\
\adjustbox{scale=1.5}{(a) SVM}
&
\adjustbox{scale=1.5}{(b) $k$-NN}
&
\adjustbox{scale=1.5}{(c) RF}
\\ \\
\input figures/random_random/conf_digram_MLP.tex
&
\input figures/random_random/conf_digram_ELM.tex
&
\input figures/random_random/conf_digram_LSTM.tex
\\
\adjustbox{scale=1.5}{(d) MLP}
&
\adjustbox{scale=1.5}{(e) ELM}
&
\adjustbox{scale=1.5}{(f) LSTM}
\\ \\
& \input figures/random_random/conf_digram_CNN.tex &
\\ \\
& \adjustbox{scale=1.5}{(g) CNN}
\end{tabular}
}
\caption{Random-random confusion matrices for digram features}\label{figure:Digram}
\end{figure}

\clearpage

\subsection*{Summary of Results for Fixed-Fixed Scenario}

\begin{figure}[!htb]
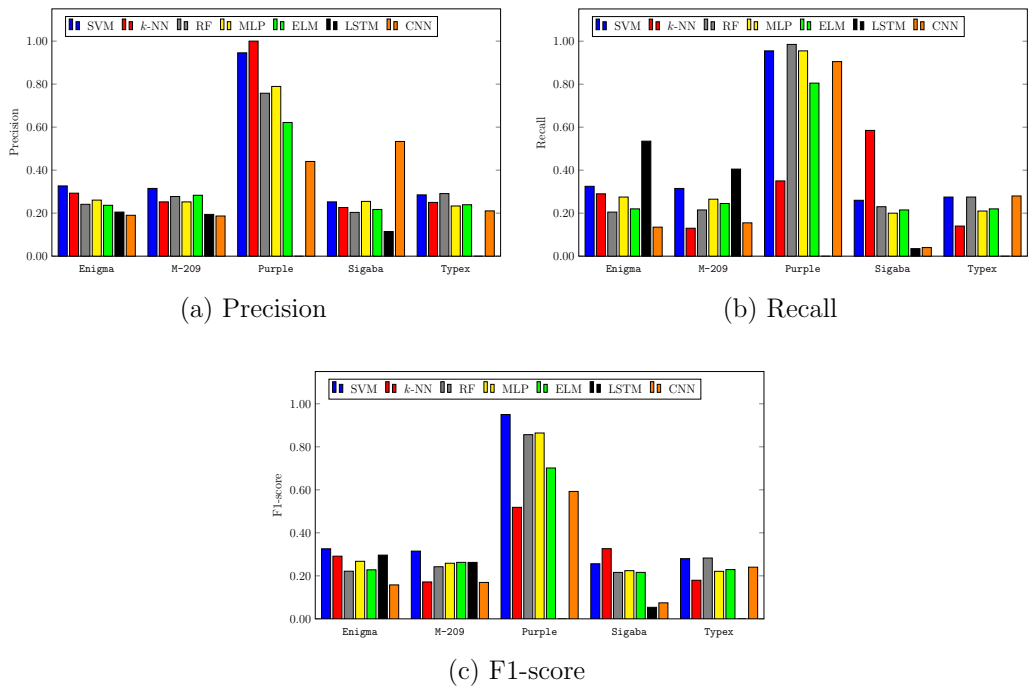

\centering
\begin{tabular}{cc}
\begin{adjustbox}{width=0.425\textwidth}
\input figures/fixed_fixed/seqPrec.tex
\end{adjustbox}
&
\begin{adjustbox}{width=0.425\textwidth}
\input figures/fixed_fixed/seqRecall.tex
\end{adjustbox}
\\
\adjustbox{scale=0.85}{(a) Precision}
&
\adjustbox{scale=0.85}{(b) Recall}
\\
\\
\multicolumn{2}{c}{
\begin{adjustbox}{width=0.425\textwidth}
\input figures/fixed_fixed/seqF1.tex
\end{adjustbox}
}
\\
\multicolumn{2}{c}{\adjustbox{scale=0.85}{(c) F1-score}}
\end{tabular} 
\caption{Fixed-fixed precision, recall, and F1-scores for letter sequence features}\label{figure:FFSeqScore}
\end{figure}

In this section, we give results for the fixed-fixed scenario,
which is discussed in Section~\ref{sect:FF}. We experiment with each of the
seven models (SVM, $k$-NN, RF, MLP, ELM, LSTM, and CNN) using 
each of the three feature types (histogram, digram, and letter sequence)
for a total of~21 distinct models. For each model, we provide a confusion
matrix, and we give the precision, recall, and F1-score
for each of the five ciphers in bar graph form.
The confusion matrices and bar graphs for the histogram
features appear in Figures~\ref{figure:FFHist} and~\ref{figure:FFHistScore},
respectively; the confusion matrices and bar graphs for the digram
features are in Figures~\ref{figure:FFDigram} and~\ref{figure:FFDigramScore}, 
respectively; and the confusion matrices and bar graphs for the letter sequence
features are in Figures~\ref{figure:FFSeq} and~\ref{figure:FFSeqScore}.

\begin{figure}[!htb]
\centering
\adjustbox{scale=0.525}{
\begin{tabular}{ccc}
\input figures/fixed_fixed/conf_hist_SVM.tex
&
\input figures/fixed_fixed/conf_hist_kNN.tex
&
\input figures/fixed_fixed/conf_hist_RF.tex
\\
\adjustbox{scale=1.5}{(a) SVM}
&
\adjustbox{scale=1.5}{(b) $k$-NN}
&
\adjustbox{scale=1.5}{(c) RF}
\\ \\
\input figures/fixed_fixed/conf_hist_MLP.tex
&
\input figures/fixed_fixed/conf_hist_ELM.tex
&
\input figures/fixed_fixed/conf_hist_LSTM.tex
\\
\adjustbox{scale=1.5}{(d) MLP}
&
\adjustbox{scale=1.5}{(e) ELM}
&
\adjustbox{scale=1.5}{(f) LSTM}
\\ \\
& \input figures/fixed_fixed/conf_hist_CNN.tex &
\\ \\
& \adjustbox{scale=1.5}{(g) CNN}
\end{tabular}
}
\caption{Fixed-fixed confusion matrices for histogram features}\label{figure:FFHist}
\end{figure}

\begin{figure}[!htb]
\centering
\begin{tabular}{cc}
\begin{adjustbox}{width=0.425\textwidth}
\input figures/fixed_fixed/histoPrec.tex
\end{adjustbox}
&
\begin{adjustbox}{width=0.425\textwidth}
\input figures/fixed_fixed/histoRecall.tex
\end{adjustbox}
\\
\adjustbox{scale=0.85}{(a) Precision}
&
\adjustbox{scale=0.85}{(b) Recall}
\\
\\
\multicolumn{2}{c}{
\begin{adjustbox}{width=0.425\textwidth}
\input figures/fixed_fixed/histoF1.tex
\end{adjustbox}
}
\\
\multicolumn{2}{c}{\adjustbox{scale=0.85}{(c) F1-score}}
\end{tabular} 
\caption{Fixed-fixed precision, recall, and F1-scores for histogram features}\label{figure:FFHistScore}
\end{figure}


\begin{figure}[!htb]
\centering
\adjustbox{scale=0.525}{
\begin{tabular}{ccc}
\input figures/fixed_fixed/conf_digram_SVM.tex
&
\input figures/fixed_fixed/conf_digram_kNN.tex
&
\input figures/fixed_fixed/conf_digram_RF.tex
\\
\adjustbox{scale=1.5}{(a) SVM}
&
\adjustbox{scale=1.5}{(b) $k$-NN}
&
\adjustbox{scale=1.5}{(c) RF}
\\ \\
\input figures/fixed_fixed/conf_digram_MLP.tex
&
\input figures/fixed_fixed/conf_digram_ELM.tex
&
\input figures/fixed_fixed/conf_digram_LSTM.tex
\\
\adjustbox{scale=1.5}{(d) MLP}
&
\adjustbox{scale=1.5}{(e) ELM}
&
\adjustbox{scale=1.5}{(f) LSTM}
\\ \\
& \input figures/fixed_fixed/conf_digram_CNN.tex &
\\ \\
& \adjustbox{scale=1.5}{(g) CNN}
\end{tabular}
}
\caption{Fixed-fixed confusion matrices for digram features}\label{figure:FFDigram}
\end{figure}

\begin{figure}[!htb]
\centering
\begin{tabular}{cc}
\begin{adjustbox}{width=0.425\textwidth}
\input figures/fixed_fixed/digramPrec.tex
\end{adjustbox}
&
\begin{adjustbox}{width=0.425\textwidth}
\input figures/fixed_fixed/digramRecall.tex
\end{adjustbox}
\\
\adjustbox{scale=0.85}{(a) Precision}
&
\adjustbox{scale=0.85}{(b) Recall}
\\
\\
\multicolumn{2}{c}{
\begin{adjustbox}{width=0.425\textwidth}
\input figures/fixed_fixed/digramF1.tex
\end{adjustbox}
}
\\
\multicolumn{2}{c}{\adjustbox{scale=0.85}{(c) F1-score}}
\end{tabular} 
\caption{Fixed-fixed precision, recall, and F1-scores for digram features}\label{figure:FFDigramScore}
\end{figure}


\begin{figure}[!htb]
\centering
\adjustbox{scale=0.525}{
\begin{tabular}{ccc}
\input figures/fixed_fixed/conf_seq_SVM.tex
&
\input figures/fixed_fixed/conf_seq_kNN.tex
&
\input figures/fixed_fixed/conf_seq_RF.tex
\\
\adjustbox{scale=1.5}{(a) SVM}
&
\adjustbox{scale=1.5}{(b) $k$-NN}
&
\adjustbox{scale=1.5}{(c) RF}
\\ \\
\input figures/fixed_fixed/conf_seq_MLP.tex
&
\input figures/fixed_fixed/conf_seq_ELM.tex
&
\input figures/fixed_fixed/conf_seq_LSTM.tex
\\
\adjustbox{scale=1.5}{(d) MLP}
&
\adjustbox{scale=1.5}{(e) ELM}
&
\adjustbox{scale=1.5}{(f) LSTM}
\\ \\
& \input figures/fixed_fixed/conf_seq_CNN.tex &
\\ \\
& \adjustbox{scale=1.5}{(g) CNN}
\end{tabular}
}
\caption{Fixed-fixed confusion matrices for letter sequence feature}\label{figure:FFSeq}
\end{figure}

\clearpage

\subsection*{Summary of Results for Random-Fixed Scenario}

\begin{figure}[!htb]
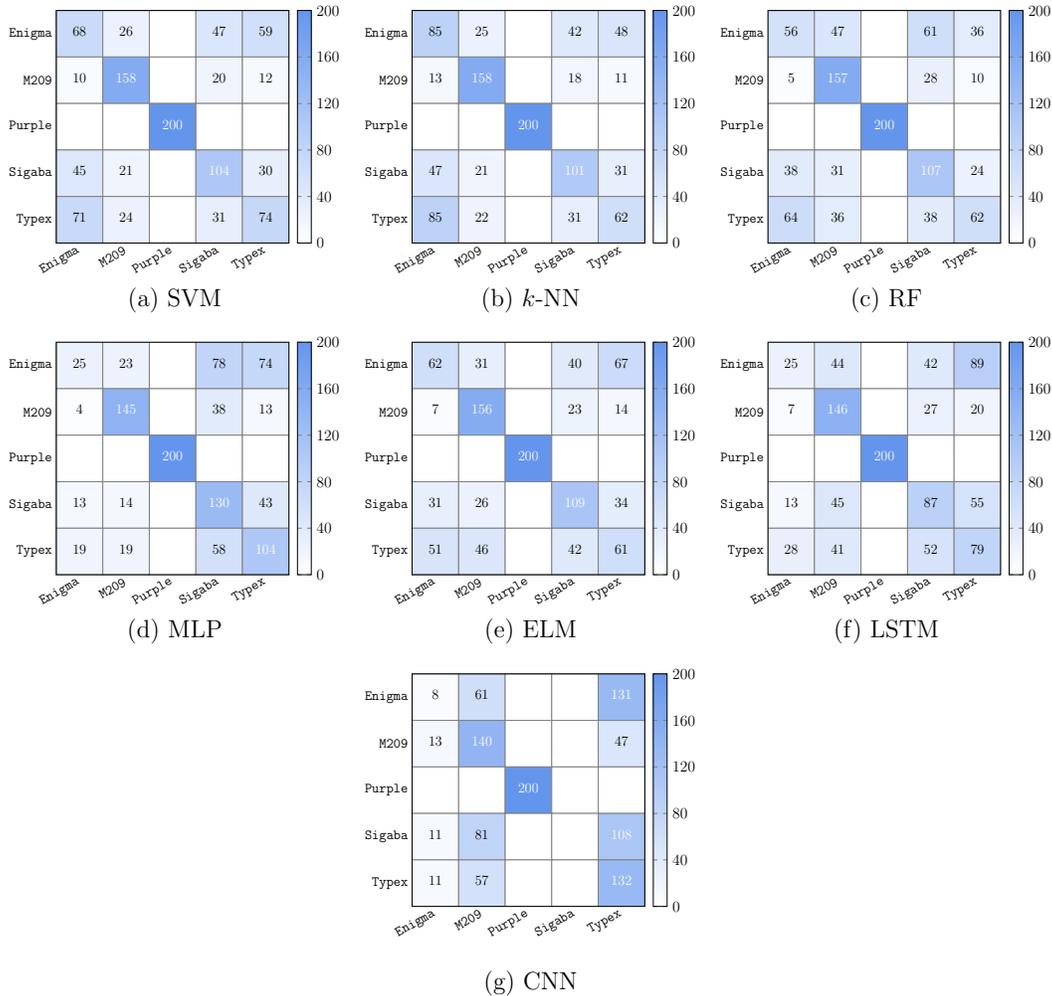

\centering
\adjustbox{scale=0.525}{
\begin{tabular}{ccc}
\input figures/random_fixed/conf_hist_SVM.tex
&
\input figures/random_fixed/conf_hist_kNN.tex
&
\input figures/random_fixed/conf_hist_RF.tex
\\
\adjustbox{scale=1.5}{(a) SVM}
&
\adjustbox{scale=1.5}{(b) $k$-NN}
&
\adjustbox{scale=1.5}{(c) RF}
\\ \\
\input figures/random_fixed/conf_hist_MLP.tex
&
\input figures/random_fixed/conf_hist_ELM.tex
&
\input figures/random_fixed/conf_hist_LSTM.tex
\\
\adjustbox{scale=1.5}{(d) MLP}
&
\adjustbox{scale=1.5}{(e) ELM}
&
\adjustbox{scale=1.5}{(f) LSTM}
\\ \\
& \input figures/random_fixed/conf_hist_CNN.tex &
\\ \\
& \adjustbox{scale=1.5}{(g) CNN}
\end{tabular}
}
\caption{Random-fixed confusion matrices for histogram features}\label{figure:RFHist}
\end{figure}

\begin{figure}[!htb]
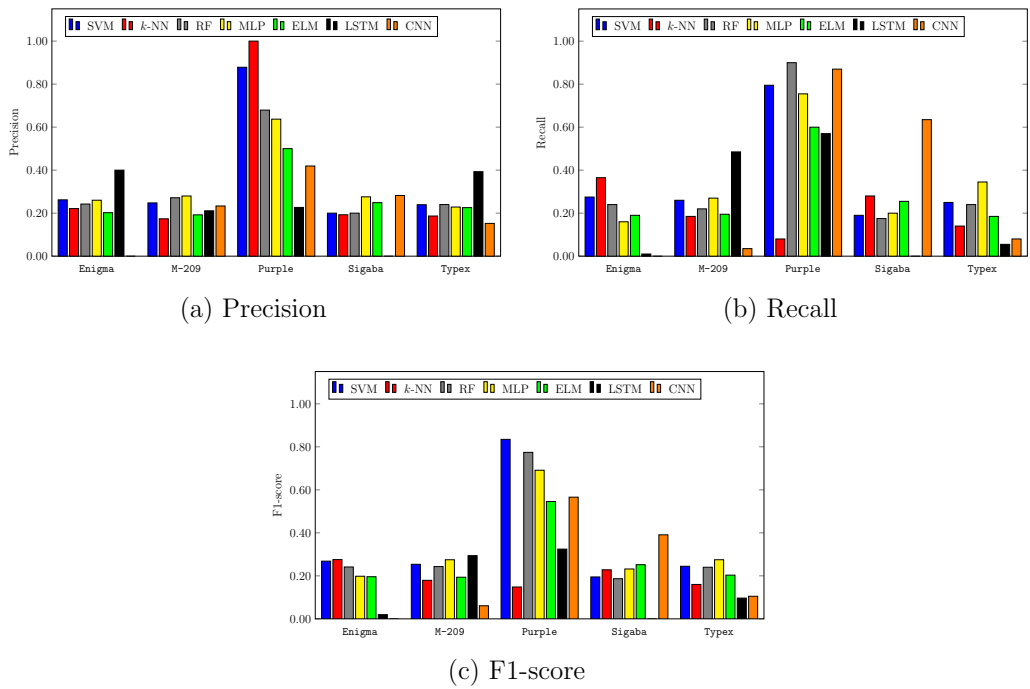

\centering
\begin{tabular}{cc}
\begin{adjustbox}{width=0.425\textwidth}
\input figures/random_fixed/seqPrec.tex
\end{adjustbox}
&
\begin{adjustbox}{width=0.425\textwidth}
\input figures/random_fixed/seqRecall.tex
\end{adjustbox}
\\
\adjustbox{scale=0.85}{(a) Precision}
&
\adjustbox{scale=0.85}{(b) Recall}
\\
\\
\multicolumn{2}{c}{
\begin{adjustbox}{width=0.425\textwidth}
\input figures/random_fixed/seqF1.tex
\end{adjustbox}
}
\\
\multicolumn{2}{c}{\adjustbox{scale=0.85}{(c) F1-score}}
\end{tabular} 
\caption{Random-fixed precision, recall, and F1-scores for letter sequence features}\label{figure:RFSeqScore}
\end{figure}

In this section, we give results for the random-fixed scenario,
which is discussed in Section~\ref{sect:RF}. We experiment with each of the
seven models (SVM, $k$-NN, RF, MLP, ELM, LSTM, and CNN) using 
each of the three feature types (histogram, digram, and letter sequence)
for a total of~21 distinct models. For each model, we provide a confusion
matrix, and we give the precision, recall, and F1-score
for each of the five ciphers in bar graph form.
The confusion matrices and bar graphs for the histogram
features appear in Figures~\ref{figure:RFHist} and~\ref{figure:RFHistScore},
respectively; the confusion matrices and bar graphs for the digram
features are in Figures~\ref{figure:RFDigram} and~\ref{figure:RFDigramScore}, 
respectively; and the confusion matrices and bar graphs for the letter sequence
features are in Figures~\ref{figure:RFSeq} and~\ref{figure:RFSeqScore}.

\begin{figure}[!htb]
\centering
\begin{tabular}{cc}
\begin{adjustbox}{width=0.425\textwidth}
\input figures/random_fixed/histoPrec.tex
\end{adjustbox}
&
\begin{adjustbox}{width=0.425\textwidth}
\input figures/random_fixed/histoRecall.tex
\end{adjustbox}
\\
\adjustbox{scale=0.85}{(a) Precision}
&
\adjustbox{scale=0.85}{(b) Recall}
\\
\\
\multicolumn{2}{c}{
\begin{adjustbox}{width=0.425\textwidth}
\input figures/random_fixed/histoF1.tex
\end{adjustbox}
}
\\
\multicolumn{2}{c}{\adjustbox{scale=0.85}{(c) F1-score}}
\end{tabular} 
\caption{Random-fixed precision, recall, and F1-scores for histogram features}\label{figure:RFHistScore}
\end{figure}


\begin{figure}[!htb]
\centering
\adjustbox{scale=0.525}{
\begin{tabular}{ccc}
\input figures/random_fixed/conf_digram_SVM.tex
&
\input figures/random_fixed/conf_digram_kNN.tex
&
\input figures/random_fixed/conf_digram_RF.tex
\\
\adjustbox{scale=1.5}{(a) SVM}
&
\adjustbox{scale=1.5}{(b) $k$-NN}
&
\adjustbox{scale=1.5}{(c) RF}
\\ \\
\input figures/random_fixed/conf_digram_MLP.tex
&
\input figures/random_fixed/conf_digram_ELM.tex
&
\input figures/random_fixed/conf_digram_LSTM.tex
\\
\adjustbox{scale=1.5}{(d) MLP}
&
\adjustbox{scale=1.5}{(e) ELM}
&
\adjustbox{scale=1.5}{(f) LSTM}
\\ \\
& \input figures/random_fixed/conf_digram_CNN.tex &
\\ \\
& \adjustbox{scale=1.5}{(g) CNN}
\end{tabular}
}
\caption{Random-fixed confusion matrices for digram features}\label{figure:RFDigram}
\end{figure}

\begin{figure}[!htb]
\centering
\begin{tabular}{cc}
\begin{adjustbox}{width=0.425\textwidth}
\input figures/random_fixed/digramPrec.tex
\end{adjustbox}
&
\begin{adjustbox}{width=0.425\textwidth}
\input figures/random_fixed/digramRecall.tex
\end{adjustbox}
\\
\adjustbox{scale=0.85}{(a) Precision}
&
\adjustbox{scale=0.85}{(b) Recall}
\\
\\
\multicolumn{2}{c}{
\begin{adjustbox}{width=0.425\textwidth}
\input figures/random_fixed/digramF1.tex
\end{adjustbox}
}
\\
\multicolumn{2}{c}{\adjustbox{scale=0.85}{(c) F1-score}}
\end{tabular} 
\caption{Random-fixed precision, recall, and F1-scores for digram features}\label{figure:RFDigramScore}
\end{figure}


\begin{figure}[!htb]
\centering
\adjustbox{scale=0.525}{
\begin{tabular}{ccc}
\input figures/random_fixed/conf_seq_SVM.tex
&
\input figures/random_fixed/conf_seq_kNN.tex
&
\input figures/random_fixed/conf_seq_RF.tex
\\
\adjustbox{scale=1.5}{(a) SVM}
&
\adjustbox{scale=1.5}{(b) $k$-NN}
&
\adjustbox{scale=1.5}{(c) RF}
\\ \\
\input figures/random_fixed/conf_seq_MLP.tex
&
\input figures/random_fixed/conf_seq_ELM.tex
&
\input figures/random_fixed/conf_seq_LSTM.tex
\\
\adjustbox{scale=1.5}{(d) MLP}
&
\adjustbox{scale=1.5}{(e) ELM}
&
\adjustbox{scale=1.5}{(f) LSTM}
\\ \\
& \input figures/random_fixed/conf_seq_CNN.tex &
\\ \\
& \adjustbox{scale=1.5}{(g) CNN}
\end{tabular}
}
\caption{Random-fixed confusion matrices for letter sequence feature}\label{figure:RFSeq}
\end{figure}

\clearpage

\subsection*{Summary of Results for Fixed-Random Scenario}

In this section, we give results for the fixed-random scenario,
which is discussed in Section~\ref{sect:FR}. We experiment with each of the
seven models (SVM, $k$-NN, RF, MLP, ELM, LSTM, and CNN) using 
each of the three feature types (histogram, digram, and letter sequence)
for a total of~21 distinct models. For each model, we provide a confusion
matrix, and we give the precision, recall, and F1-score
for each of the five ciphers in bar graph form.
The confusion matrices and bar graphs for the histogram
features appear in Figures~\ref{figure:FRHist} and~\ref{figure:FRHistScore},
respectively; the confusion matrices and bar graphs for the digram
features are in Figures~\ref{figure:FRDigram} and~\ref{figure:FRDigramScore}, 
respectively; and the confusion matrices and bar graphs for the letter sequence
features are in Figures~\ref{figure:FRSeq} and~\ref{figure:FRSeqScore}.


\begin{figure}[!htb]
\centering
\adjustbox{scale=0.525}{
\begin{tabular}{ccc}
\input figures/fixed_random/conf_hist_SVM.tex
&
\input figures/fixed_random/conf_hist_kNN.tex
&
\input figures/fixed_random/conf_hist_RF.tex
\\
\adjustbox{scale=1.5}{(a) SVM}
&
\adjustbox{scale=1.5}{(b) $k$-NN}
&
\adjustbox{scale=1.5}{(c) RF}
\\ \\
\input figures/fixed_random/conf_hist_MLP.tex
&
\input figures/fixed_random/conf_hist_ELM.tex
&
\input figures/fixed_random/conf_hist_LSTM.tex
\\
\adjustbox{scale=1.5}{(d) MLP}
&
\adjustbox{scale=1.5}{(e) ELM}
&
\adjustbox{scale=1.5}{(f) LSTM}
\\ \\
& \input figures/fixed_random/conf_hist_CNN.tex &
\\ \\
& \adjustbox{scale=1.5}{(g) CNN}
\end{tabular}
}
\caption{Fixed-random confusion matrices for histogram features}\label{figure:FRHist}
\end{figure}

\begin{figure}[!htb]
\centering
\begin{tabular}{cc}
\begin{adjustbox}{width=0.425\textwidth}
\input figures/fixed_random/histoPrec.tex
\end{adjustbox}
&
\begin{adjustbox}{width=0.425\textwidth}
\input figures/fixed_random/histoRecall.tex
\end{adjustbox}
\\
\adjustbox{scale=0.85}{(a) Precision}
&
\adjustbox{scale=0.85}{(b) Recall}
\\
\\
\multicolumn{2}{c}{
\begin{adjustbox}{width=0.425\textwidth}
\input figures/fixed_random/histoF1.tex
\end{adjustbox}
}
\\
\multicolumn{2}{c}{\adjustbox{scale=0.85}{(c) F1-score}}
\end{tabular} 
\caption{Fixed-random precision, recall, and F1-scores for histogram features}\label{figure:FRHistScore}
\end{figure}


\begin{figure}[!htb]
\centering
\adjustbox{scale=0.525}{
\begin{tabular}{ccc}
\input figures/fixed_random/conf_digram_SVM.tex
&
\input figures/fixed_random/conf_digram_kNN.tex
&
\input figures/fixed_random/conf_digram_RF.tex
\\
\adjustbox{scale=1.5}{(a) SVM}
&
\adjustbox{scale=1.5}{(b) $k$-NN}
&
\adjustbox{scale=1.5}{(c) RF}
\\ \\
\input figures/fixed_random/conf_digram_MLP.tex
&
\input figures/fixed_random/conf_digram_ELM.tex
&
\input figures/fixed_random/conf_digram_LSTM.tex
\\
\adjustbox{scale=1.5}{(d) MLP}
&
\adjustbox{scale=1.5}{(e) ELM}
&
\adjustbox{scale=1.5}{(f) LSTM}
\\ \\
& \input figures/fixed_random/conf_digram_CNN.tex &
\\ \\
& \adjustbox{scale=1.5}{(g) CNN}
\end{tabular}
}
\caption{Fixed-random confusion matrices for digram features}\label{figure:FRDigram}
\end{figure}

\begin{figure}[!htb]
\centering
\begin{tabular}{cc}
\begin{adjustbox}{width=0.425\textwidth}
\input figures/fixed_random/digramPrec.tex
\end{adjustbox}
&
\begin{adjustbox}{width=0.425\textwidth}
\input figures/fixed_random/digramRecall.tex
\end{adjustbox}
\\
\adjustbox{scale=0.85}{(a) Precision}
&
\adjustbox{scale=0.85}{(b) Recall}
\\
\\
\multicolumn{2}{c}{
\begin{adjustbox}{width=0.425\textwidth}
\input figures/fixed_random/digramF1.tex
\end{adjustbox}
}
\\
\multicolumn{2}{c}{\adjustbox{scale=0.85}{(c) F1-score}}
\end{tabular} 
\caption{Fixed-random precision, recall, and F1-scores for digram features}\label{figure:FRDigramScore}
\end{figure}


\begin{figure}[!htb]
\centering
\adjustbox{scale=0.525}{
\begin{tabular}{ccc}
\input figures/fixed_random/conf_seq_SVM.tex
&
\input figures/fixed_random/conf_seq_kNN.tex
&
\input figures/fixed_random/conf_seq_RF.tex
\\
\adjustbox{scale=1.5}{(a) SVM}
&
\adjustbox{scale=1.5}{(b) $k$-NN}
&
\adjustbox{scale=1.5}{(c) RF}
\\ \\
\input figures/fixed_random/conf_seq_MLP.tex
&
\input figures/fixed_random/conf_seq_ELM.tex
&
\input figures/fixed_random/conf_seq_LSTM.tex
\\
\adjustbox{scale=1.5}{(d) MLP}
&
\adjustbox{scale=1.5}{(e) ELM}
&
\adjustbox{scale=1.5}{(f) LSTM}
\\ \\
& \input figures/fixed_random/conf_seq_CNN.tex &
\\ \\
& \adjustbox{scale=1.5}{(g) CNN}
\end{tabular}
}
\caption{Fixed-random confusion matrices for letter sequence feature}\label{figure:FRSeq}
\end{figure}

\begin{figure}[!htb]
\centering
\begin{tabular}{cc}
\begin{adjustbox}{width=0.425\textwidth}
\input figures/fixed_random/seqPrec.tex
\end{adjustbox}
&
\begin{adjustbox}{width=0.425\textwidth}
\input figures/fixed_random/seqRecall.tex
\end{adjustbox}
\\
\adjustbox{scale=0.85}{(a) Precision}
&
\adjustbox{scale=0.85}{(b) Recall}
\\
\\
\multicolumn{2}{c}{
\begin{adjustbox}{width=0.425\textwidth}
\input figures/fixed_random/seqF1.tex
\end{adjustbox}
}
\\
\multicolumn{2}{c}{\adjustbox{scale=0.85}{(c) F1-score}}
\end{tabular} 
\caption{Fixed-random precision, recall, and F1-scores for letter sequence features}\label{figure:FRSeqScore}
\end{figure}



\end{document}